\documentclass{article}

    \PassOptionsToPackage{numbers,sort&compress}{natbib}

    \usepackage[preprint]{neurips_2020}

\usepackage[utf8]{inputenc} 
\usepackage[T1]{fontenc}    
\usepackage[colorlinks]{hyperref}       
\usepackage{url}            
\usepackage{booktabs}       
\usepackage{amsfonts}       
\usepackage{nicefrac}       
\usepackage{microtype}      
\usepackage{graphicx}
\usepackage{amsmath}
\usepackage{subcaption}
\usepackage{textcomp}
\usepackage{multicol}
\usepackage{multirow}
\newcommand{\rarrow}{$\rightarrow$}

\usepackage{xcolor}
\usepackage{paralist}

\makeatletter
\newcommand{\settitle}{\@maketitle}
\makeatother

\def\papertitle{AVLnet: Learning Audio-Visual Language Representations from Instructional Videos}
\title{\papertitle}

\author{%
    \textbf{Andrew Rouditchenko$^1$}\thanks{These authors contributed equally to this work} \enskip
    \textbf{Angie Boggust$^1$}\footnotemark[1] \enskip
    \textbf{David Harwath$^2$} \enskip
    \textbf{Brian Chen$^{3}$} \vspace{1mm} \\
    \textbf{Dhiraj Joshi$^{4}$} \enskip
    \textbf{Samuel Thomas$^{4}$} \enskip
    \textbf{Kartik Audhkhasi$^{5}$}\enskip
    \textbf{Hilde Kuehne}$^{4}$ \enskip
    \textbf{Rameswar Panda}$^{4}$ \vspace{1mm} \\ 
    \textbf{Rogerio Feris}$^{4}$ \enskip
    \textbf{Brian Kingsbury$^{4}$} \enskip
    \textbf{Michael Picheny$^{6}$} \enskip
    \textbf{Antonio Torralba$^1$} \enskip
    \textbf{James Glass$^1$}
    \vspace{1mm} \\
    $^1$MIT CSAIL, $^2$UT Austin, $^3$Columbia University, $^4$IBM Research AI, $^5$Google, $^6$NYU 
    \vspace{1mm} \\
    \small{
        \texttt{roudi@mit.edu}
    }
}

\begin{document}

\maketitle

\begin{abstract}
Current methods for learning visually grounded language from videos often rely on text annotation, such as human generated captions or machine generated automatic speech recognition (ASR) transcripts.
In this work, we introduce the Audio-Video Language Network (AVLnet), a self-supervised network that learns a shared audio-visual embedding space directly from raw video inputs.
To circumvent the need for text annotation, we learn audio-visual representations from randomly segmented video clips and their raw audio waveforms.
We train AVLnet on HowTo100M, a large corpus of publicly available instructional videos, and evaluate on image retrieval and video retrieval tasks, achieving state-of-the-art performance.
We perform analysis of AVLnet's learned representations, showing our model utilizes speech and natural sounds to learn audio-visual concepts.
Further, we propose a tri-modal model that jointly processes raw audio, video, and text captions from videos to learn a multi-modal semantic embedding space useful for text-video retrieval.
Our code, data, and trained models will be released at \href{http://avlnet.csail.mit.edu/}{avlnet.csail.mit.edu}.
\end{abstract}
\section{Introduction}
\label{sec:introduction}
Humans learn to understand language, recognize objects, and identify correspondences between the two by recognizing patterns in what they see and what they hear.
Researchers have developed machine learning models similarly capable of relating spoken words to semantically relevant images~\cite{synnaeve2014learning,harwath2016unsupervised, harwath2018jointly,harwath2020jointly,ilharco2019large,chrupala2017representations,merkx2019language,sanabria2021talk}.
By training models to retrieve images from associated spoken captions, they learn to identify words in speech and objects in images without supervised speech recognition or object detection.
However, these methods require the collection of recorded spoken captions, limiting their scalability to other languages and visual contexts.

Videos provide a natural source of paired visual and audio data that does not require manual annotation and exists publicly in large quantities.
Thus, self-supervised audio-video models~\cite{aytar2016soundnet,arandjelovic2017look,arandjelovic2018objects,owens2018audio,korbar2018cooperative, zhao2018sound,rouditchenko2019self} have been applied to cross-modal tasks focused on identifying non-speech sounds and localizing the objects that produced them.
We instead focus on relating spoken words to visual entities in videos such as objects and actions, which is a challenging task since human speech is semantically complex and the objects of interest do not produce the sound.
Towards this goal, we use instructional videos which provide opportunities to learn semantic relationships between raw speech and visual entities given the narration naturally present in them. 

A common approach for learning from instructional videos is to develop text-video models that learn a multi-modal embedding space.
These models often do not incorporate the audio signal, but even models that do~\cite{miech2018learning,mithun2018learning,wray2019fine,yu2018joint,liu2019use,holzenberger2019learning,alayrac2020self} still require text captions.
To collect captions, some methods rely on humans to generate visual descriptions~\cite{zhou2018towards}.
Unlike raw audio which can be noisy and nondescript, human-generated text provides a clean, visually salient signal; however, collecting text descriptions is time-consuming and infeasible for large datasets.
To reduce the need for annotation, other methods rely on ASR transcripts to provide text representative of the speech in videos~\cite{miech2020end,miech2019howto100m,sun2019learning,sun2019videobert,sanabria18how2}.
However, ASR transcripts process the continuous speech signal into discrete words, which limits words to a certain vocabulary and misses the opportunity to learn from visually relevant non-speech sounds.
Further, ASR can be errorful when confronted with background sounds, reverberation, and accents, all of which are found in instructional videos.
Models trained on ASR transcripts are also inapplicable to the 98\% of languages for which ASR is unavailable~\cite{prasad2019building}. 
For these reasons, our goal is to learn from the raw audio and visual channels in videos without any additional annotation or ASR transcripts.

In response, we propose the Audio-Video Language Network (AVLnet) and a self-supervised framework to learn visually grounded language from raw video input.
We circumvent the need for spoken or textual annotations by learning directly from the raw audio channel in video clips.
Our model consists of audio and video branches that extract local video clip features and pool them into single feature vectors representing the content in each modality.
We apply non-linear feature gating~\cite{miech2017learnable} enabling our model to re-calibrate the feature activations before the final output embeddings.
To train our model on the noisy audio signal in instructional videos, we utilize the Masked Margin Softmax (MMS) loss~\cite{ilharco2019large} to simulate audio and visual retrieval and robustly train against a large number of negative samples.
This results in an audio-video embedding space that colocates semantically similar audio and visual inputs and can successfully be used for downstream retrieval tasks.

\begin{figure*}[t!]
    \centering
    \includegraphics[width=\textwidth]{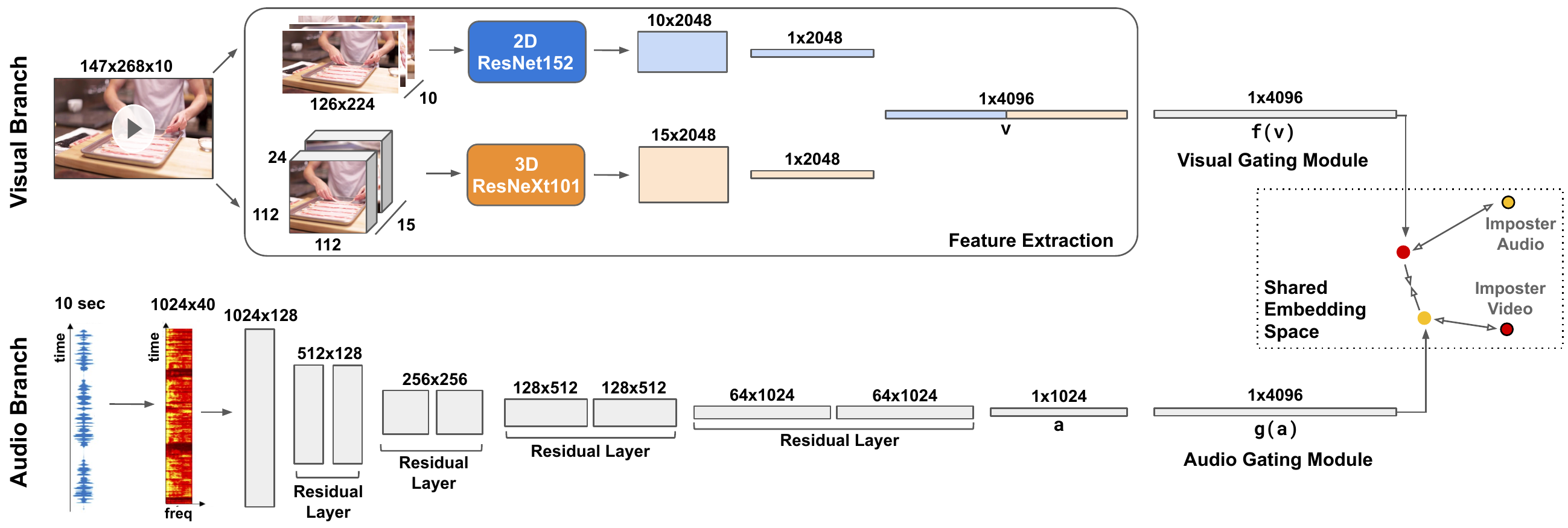}
    \captionof{figure}{The Audio-Video Language Network (AVLnet) model consists of video and audio branches, non-linear feature gating, and an audio-video embedding space. The model is trained through self-supervision and applied to image and video retrieval tasks.}
    \label{fig:avlnet}
\end{figure*}

We train AVLnet on HowTo100M~\cite{miech2019howto100m}, a large-scale instructional video dataset.
Instead of defining video clips at ASR boundaries, we train our model on randomly segmented clips, reducing the need for supervision. 
Despite training on unlabeled videos, our model achieves state-of-the-art retrieval results on speech-image pairs in the Places Audio Caption dataset~\cite{harwath2018jointly}. 
We propose video retrieval tasks on three video datasets, YouCook2~\cite{zhou2018towards}, CrossTask~\cite{zhukov2019cross}, and MSR-VTT~\cite{xu2016msr-vtt}.
We further show how our model leverages audio cues from both speech and natural sounds for retrieval and semantically relates the audio and visual modalities to learn audio-visual concepts.

Learning without text captions is desirable since ASR is only supported for less than $2\%$ of the world's spoken languages and manually annotating videos with captions is expensive and time-consuming.
However, many existing video datasets already have text captions.
Therefore, we also introduce a text branch into the AVLnet model to process text.
We refer to the resulting class of models as AVLnet-Text.
We propose two ways to incorporate the text branch into the AVLnet model with two corresponding training losses.
We compare our approach with previous text-video models on several standard video and language datasets: YouCook2~\cite{zhou2018towards}, MSR-VTT~\cite{xu2016msr-vtt}, and LSMDC~\cite{rohrbach2017movie}.
Finally, we show that AVLnet trained without text captions on HowTo100M can perform retrieval with text on the downstream datasets in both the zero-shot and fine-tuned settings, which suggests that the audio representations can be adapted with text representations from only a small amount of text captions.

\section{Related Work}
\textbf{Learning Visually-Grounded Speech.}
The task of matching spoken audio captions to semantically relevant images was introduced in the effort to build models that learn language from raw audio and visual semantic supervision~\cite{synnaeve2014learning,harwath2015deep,harwath2016unsupervised}.
Models are typically trained to learn an audio-visual embedding space where true image-caption pairs are similar to each other, while non-matching pairs are far apart.
Over the years, researchers have proposed modeling improvements with more complex image encoders, audio encoders, and loss functions~\cite{harwath2018jointly,harwath2020learning,harwath2020jointly,chrupala2017representations,merkx2019language,ilharco2019large,suris2019learning,mortazavi2020speech,sanabria2021talk,wang2021align}.
In terms of training data, Harwath~et~al.~\cite{harwath2016unsupervised,harwath2018jointly,harwath2020jointly} collected 400k spoken audio captions of images in the Places205~\cite{zhou2014learning} dataset in English from 2,683 speakers, which is one of the largest spoken caption datasets.
Other work has proposed synthetic speech captions as training data, which are less natural~\cite{chrupala2017representations,havard2017speech,ilharco2019large}.
The models have been explored for other tasks such as speech retrieval given spoken queries or text captions~\cite{kamper2018visually,kamper2019semantic}, discovering word-like speech units~\cite{harwath2017learning,wang2019multimodal,wang2020dnn}, and for other data such as handwritten digits and spoken captions~\cite{leidal2017learning,eloff2019multimodal,hsu2018disentangling}.
For a recent survey of visually grounded models of spoken language, see Chrupała~\cite{chrupala2021visually}.
We instead use videos naturally present on the internet as the primary source of training data, which are available in English and in other languages.
While we focus on the spoken narration naturally present in instructional videos, researchers have collected spoken captions for videos~\cite{monfort2021spoken,oncescu2020queryd} in concurrent work.

\textbf{Self-Supervised Audio-Video Learning.}
Self-supervised audio-video learning has been explored in recent years to learn representations of objects and sounds without manually labelled data.
Some works propose proxy tasks to learn representations for downstream tasks such as classification~\cite{aytar2016soundnet,arandjelovic2017look,korbar2018cooperative,owens2016ambient,owens2018audio,hu2019deep,alwassel_2020_xdc}.
Other approaches use self-supervised learning for audio-video applications, such as audio-visual source separation~\cite{gao2018learning,zhao2018sound,rouditchenko2019self} and spatial audio generation~\cite{gao20192,morgado2018self,yang2020telling}.
The most relevant works are those that apply audio-video models for cross-modal retrieval tasks. 
Boggust~et~al.~\cite{boggust2019grounding} aimed to reduce the amount of annotation required for image and spoken caption models and instead used videos as training data.
They directly applied the speech-image architecture from Harwath~et~al.~\cite{harwath2018jointly} on still image frames and surrounding audio in videos.
During inference, their model samples a single image frame from video clips and performs image to audio retrieval.
We expand upon this work by developing the AVLnet architecture which learns from entire videos clips and performs video to audio retrieval. 
Arandjelovi{\'c} and Zisserman~\cite{arandjelovic2018objects} employ self-supervision between the audio and visual streams in video to relate objects with the sounds they make.
They train their model for binary classification of true audio and video pairs versus mismatched ones and apply their model for audio-video retrieval.
In our work, we instead use audio-video self-supervision to relate objects to the speech that describes them and directly train our model for audio-video retrieval.

\textbf{Multi-Modal Learning from Instructional Videos.}
The recent influx of instructional video datasets such as How2~\cite{sanabria18how2}, Inria Instructional Videos~\cite{alayrac2016unsupervised}, COIN~\cite{Tang2019COIN}, CrossTask~\cite{zhukov2019cross}, YouCook2~\cite{zhou2018towards}, Mining YouTube~\cite{kuehne2019mining}, and HowTo100M~\cite{miech2019howto100m} has inspired a variety of methods for semi-supervised text-video modelling.
These works focus on learning a joint multi-modal embedding space between text and video, and typically do not incorporate the audio signal.
Methods that do incorporate audio~\cite{liu2019use,miech2018learning,mithun2018learning,wray2019fine,yu2018joint,holzenberger2019learning,alayrac2020self} still require text captions and do not learn from the raw videos alone.
We build upon these works by learning a joint embedding space directly between video and the audio naturally present in videos, and showing that our method can also incorporate ASR text and annotated text captions when available.

\section{Technical Approach for Audio-Video Models}
\label{sec:avlnet}

\subsection{Audio-Video Models}
The AVLnet architecture (Figure~\ref{fig:avlnet}) consists of parallel visual and audio branches that extract features at a local level and then pool them into visual and audio feature vectors representing the overall content within each modality.
This procedure provides flexibility by allowing the model to handle variable length video clips, which is especially useful during inference where clip boundaries are determined by human annotators and can vary drastically in length. 
The visual branch consists of a 2D and 3D CNN feature extraction pipeline.
From each video clip, we compute 2D image features to obtain 1 feature per second using a ResNet-152 model~\cite{he2016deep} pretrained on ImageNet~\cite{deng2009imagenet} and 3D video features to obtain 1.5 features per second using a ResNeXt-101 model~\cite{hara2018can} pretrained on Kinetics~\cite{carreira2017quo}. 
Each of the CNN outputs are temporally max-pooled to produce two 2048-dimensional feature vectors, which are then concatenated into a 4096-dimensional feature vector $\textbf{v}$.
The audio branch consists of a trainable CNN with residual layers~\cite{harwath2018jointly} to process the raw audio in videos.
The model takes in audio spectrograms and outputs a temporal feature map, which is temporally mean-pooled to obtain a 1024-dimensional feature vector $\textbf{a}$.
In contrast to text-video models that require pretrained word embeddings to process speech transcripts~\cite{miech2020end,miech2019howto100m}, our audio model is not pretrained, so it can be applied to videos in any language, including those for which ASR is not available.

\subsection{Audio-Video Gated Embeddings}
After the visual feature vector $\textbf{v}$ and audio feature vector $\textbf{a}$ are extracted, we learn a projection of both vectors into a shared embedding space.
While this could be achieved with a linear projection, we apply non-linear feature gating~\cite{miech2017learnable} which allows the model to re-calibrate each dimension based on its learned importance and encourages the model to activate dimensions in unison across both modalities. 

Non-linear gating is defined as:
\begin{align}
    \label{eq:gating-visuals}
    f(\textbf{v}) &= (W^v_1 \textbf{v} + b^v_1) \circ \sigma(W^v_2 (W^v_1 \textbf{v} + b^v_1) + b^v_2) \\
    \label{eq:gating-audio}
    g(\textbf{a}) &= (W^a_1 \textbf{a} + b^a_1) \circ \sigma(W^a_2 (W^a_1 \textbf{a} + b^a_1) + b^a_2)
\end{align}
where $f(\textbf{v})$ and $g(\textbf{a})$ are the output 4096-dimensional embedding vectors, $W^a_1, W^a_2, W^v_1, W^v_2$ matrices and $b^a_1, b^a_2, b^v_1, b^v_2$ vectors are learnable parameters, $\circ$ 
denotes element-wise multiplication, and $\sigma$ is an element-wise sigmoid activation.

\subsection{Contrastive Loss for Audio-Video Retrieval}
\label{sec:loss}
Due to the self-supervised nature of AVLnet, we use the Masked Margin Softmax (MMS) loss~\cite{ilharco2019large}, a contrastive loss function that simulates retrieval within each batch.
The MMS loss trains the model to discriminate between the true audio-visual embedding pairs ($\textbf{a}_i$, $\textbf{v}_i$), and imposter pairs ($\textbf{a}_i$, $\textbf{v}_j^{\text{imp}}$) and ($\textbf{a}_j^{\text{imp}}$,~$\textbf{v}_i$).
The indices ($i$, $j$) indicate the index of the video clip in the batch.
Unlike the triplet loss used in prior unsupervised audio-image modeling~\cite{harwath2018jointly} that samples imposter pairs randomly or via negative mining, the MMS loss enables comparisons of positives with a wider range of negatives.
While the original MMS loss includes a masking component to handle multiple ground truth audio captions paired with each visual sample, we exclude the masking since it is inapplicable to our scenario where each visual clip contains only one ground truth audio pair.
The loss $\mathcal{L}_{MMS}$ is defined as follows:
\begin{equation}
\label{eq:mms}
    \mathcal{L}_{MMS}(f(\textbf{v}), g(\textbf{a})) = L(f(\textbf{v}), g(\textbf{a})) + L(g(\textbf{a}), f(\textbf{v}))
\end{equation}
Where $f(\textbf{v})$ and $g(\textbf{a})$ are the gated embeddings, and the function $L$ defined as:
\begin{equation}
  \label{eq:softmax}
    \resizebox{.5\linewidth}{!}{$
    L(\textbf{x}, \textbf{y}) = -\frac{1}{B} \sum\limits_{i=1}^{B} \Bigg( \log \frac{\displaystyle e^{\textbf{x}_i \cdot \textbf{y}_i - \delta}}{\displaystyle e^{\textbf{x}_i \cdot \textbf{y}_i  - \delta} + \textstyle \sum\limits_{\substack{j=1 \\ j\neq i}}^{B}e^{\textbf{x}_i \cdot \textbf{y}_j^{\text{imp}}}}\Bigg)$}
\end{equation}
The MMS loss $\mathcal{L}_{MMS}$ can be seen as the sum of two applications of InfoNCE~\cite{oord2018representation} (with a margin), the first where the visual input is fixed and audio samples are retrieved, and the second where the audio input is fixed and visual samples are retrieved.
However, whereas negatives are sampled from within the same audio sample for InfoNCE~\cite{oord2018representation}, we use audio and video samples from both within the same video and from others as negatives as this has been empirically shown to improve performance for text-video approaches~\cite{miech2019howto100m}.
During training, we use a batch of $N$ videos and sample $M$ clips per video, resulting in effective batch of $B=NM$ video clips.
An illustration of the loss is provided in Section~\ref{sec:appendix-mms} of the Appendix.

\subsection{Video Clip Sampling} 
Given a corpus of unlabeled instructional videos, we generate training samples by randomly segmenting each video into $M$ clips of length $t$ (which may overlap) to obtain a corpus of clips.
This procedure allows us to sample clips without supervised annotation (i.e., segmenting based on ASR transcripts.)
As a result, it is applicable to instructional videos in languages not supported by ASR, and it enables greater flexibility to vary the number and length of clips in the resulting dataset.
Although unsupervised clip selection may result in silent or non-salient clips, our experimental results (Section~\ref{sec:regression-analysis}) show our model performs comparably whether trained on randomly sampled clips or on clips determined by ASR boundaries.

\section{Technical Approach for Text-Audio-Video Models}

\begin{figure}[t]
    \centering 
    \begin{subfigure}{\linewidth}
        \includegraphics[width=\textwidth]{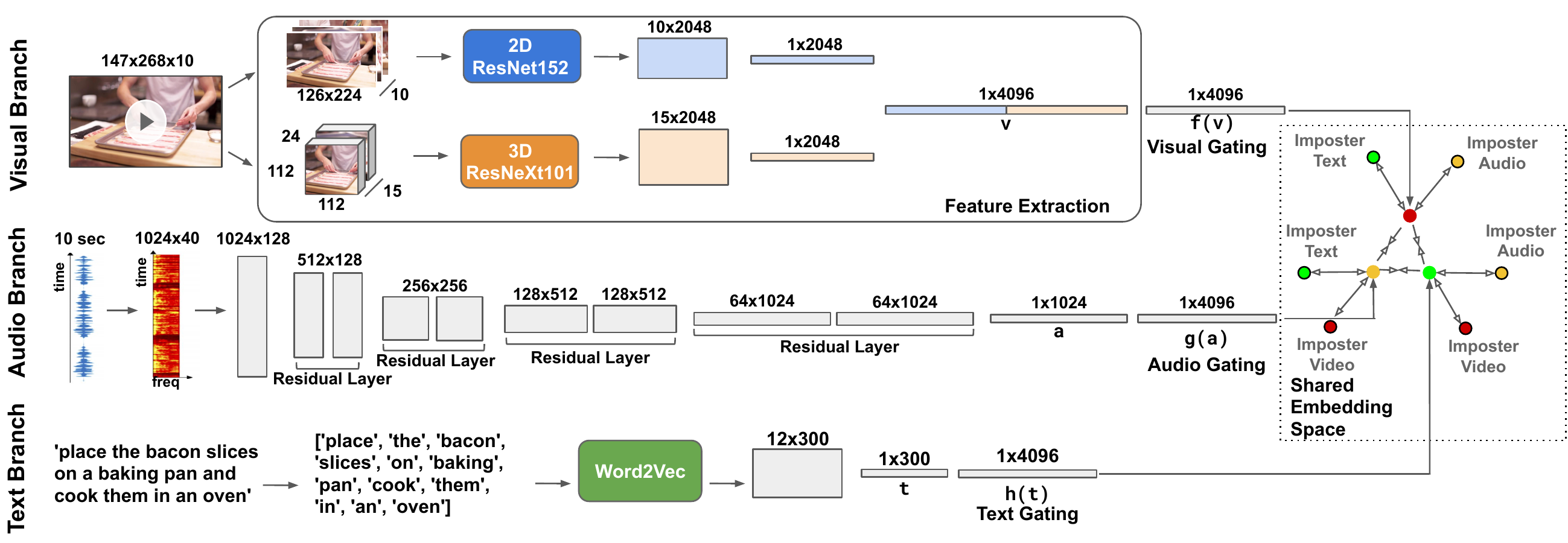}
        \captionof{figure}{AVLnet-Text-Tri Architecture}
    \end{subfigure}
    \vfill
    \vspace{0.2cm}
    \begin{subfigure}{\linewidth}
        \includegraphics[width=\textwidth]{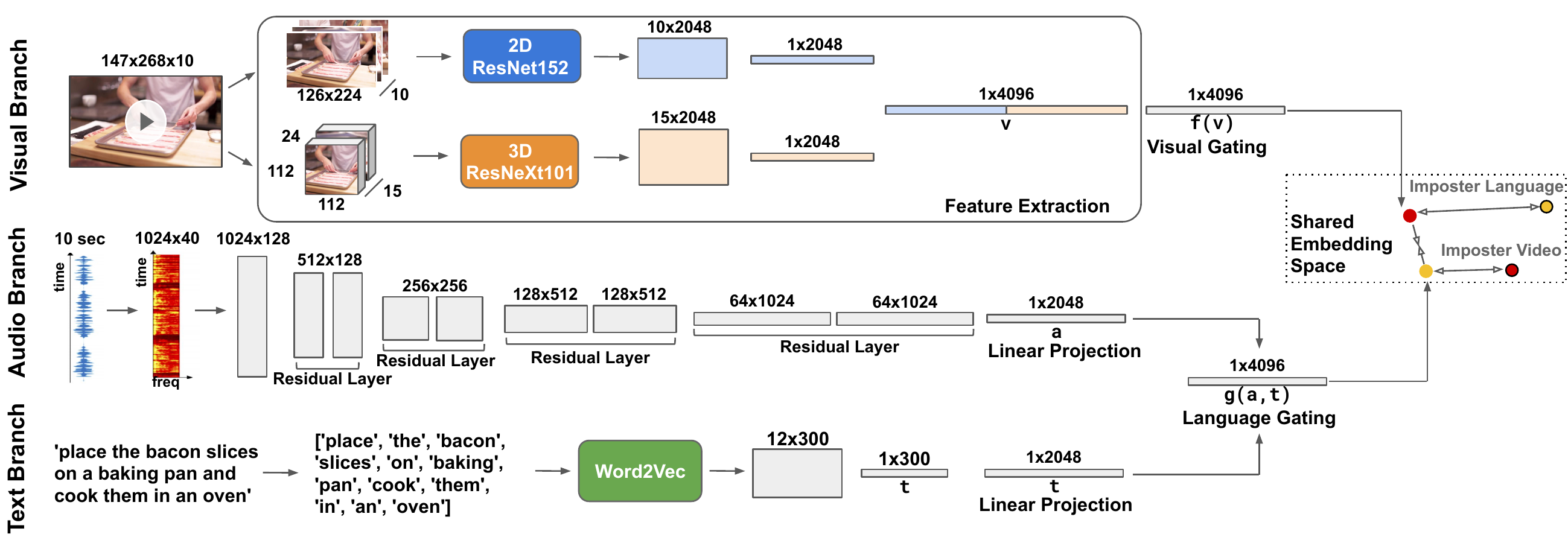}
        \captionof{figure}{AVLnet-Text-Fused Architecture}
    \end{subfigure}
    \caption{We integrate text into the AVLnet model in two different ways. The AVLnet-Text-Tri architecture keeps the text branch separate and projects all three modalities into a shared embedding space. The AVLnet-Text-Fused architecture fuses the audio and text branches into a language branch to learn a shared embedding space between the visual and language (audio and text) modalities.}
    \label{fig:avlnet_text}
\end{figure}

\subsection{Text Processing}
To incorporate text into AVLnet, we add a third branch that processes the text caption from each video clip.
We first extract word embeddings using a GoogleNews pretrained Word2Vec model~\cite{mikolov2013efficient} from a text feature extraction pipeline~\cite{miech2019howto100m}.
Following the design of the AVLnet audio and video branches, the word embeddings are max-pooled over the words in each clip's text caption.
We integrate the resulting text embedding vector into AVLnet via two different architectures.
As shown in Figure~\ref{fig:avlnet_text}, the first architecture (see Section~\ref{sec:AVLnet-Text-Tri}) keeps the text branch separate, while the second architecture (see Section~\ref{sec:AVLnet-Text-Fused}) fuses the audio and text branches.
Although our text model is shallower than recent transformer architectures, a study of deeper text models for learning a text-video embedding found little improvement over this simple text model~\cite{miech2020end}.

\subsection{Independent Tri-Modal Branch Architecture}
\label{sec:AVLnet-Text-Tri}
In this architecture, which we denote as AVLnet-Text-Tri, we keep the text, audio, and video branches separate and apply gating to each branch independently.
The motivation for this architecture is to learn a shared embedding space where any two modalities can be compared.
For a given clip, we apply non-linear gating to the max-pooled word embedding vector $\textbf{t}$ as follows:
\begin{align}
    \label{eq:gating-text}
    h(\textbf{t}) &= (W^a_1 \textbf{t} + b^t_1) \circ \sigma(W^t_2 (W^t_1 \textbf{t} + b^t_1) + b^t_2)
\end{align}
Where $h(\textbf{t})$ is the output 4096-dimensional embedding vector, $W^t_1, W^t_2$ matrices and $b^{t}_1, b^{t}_2$ vectors are learnable parameters, $\circ$ denotes element-wise multiplication, and $\sigma$ is the element-wise sigmoid activation.
We apply the MMS loss over each of the modality pairs (audio-video, audio-text, and video-text), and the branches are jointly optimized through the sum of these three losses, as follows:
\begin{equation}
    \label{eq:loss-tri}
    \mathcal{L}_{TRI}(f(\textbf{v}), g(\textbf{a}), h(\textbf{t})) = \mathcal{L}_{MMS}(f(\textbf{v}), g(\textbf{a})) + 
    \mathcal{L}_{MMS}(g(\textbf{a}), h(\textbf{t})) +
    \mathcal{L}_{MMS}(f(\textbf{v}), h(\textbf{t})) \\
\end{equation}
where $\mathcal{L}_{MMS}$ is defined in Equation~\ref{eq:mms}.

\subsection{Audio-Text Fused Architecture}
\label{sec:AVLnet-Text-Fused}
In this architecture, which we denote as AVLnet-Text-Fused, we fuse the outputs of the audio and text branches before non-linear gating due to the complementary language information in the raw audio and text.
Specifically, instead of applying the non-linear gating solely to the audio embedding vector (as in Equation~\ref{eq:gating-audio}), we apply the gating to both the audio and text embedding vectors as follows:
\begin{align}
    \label{eq:gating-text-fused}
    g(\textbf{a}, \textbf{t}) &= (W^a_1 \textbf{a} + W^t_1 \textbf{t} + b^{a+t}_1) \circ \sigma(W^{a+t}_1 (W^a_1 \textbf{a} + W^t_1 \textbf{t} + b^{a+t}_1) + b^{a+t}_2)
\end{align}
where $g(\textbf{a}, \textbf{t})$ represents the output language embedding vector combining speech and text information, $W^a_1, W^t_1, W^{a+t}_1$ matrices and $b^{a+t}_1, b^{a+t}_2$ vectors are learnable parameters, $\circ$ denotes element-wise multiplication, and $\sigma$ is the element-wise sigmoid activation.
To train this model, we optimize the following loss:
\begin{equation}
    \label{eq:loss-fused}
    \mathcal{L}_{FUSED}(f(\textbf{v}), g(\textbf{a}, \textbf{t})) = \mathcal{L}_{MMS}(f(\textbf{v}), g(\textbf{a}, \textbf{t}))
\end{equation}
where $\mathcal{L}_{MMS}$ is defined in Equation~\ref{eq:mms}.
The audio sample and text caption from each video clip are treated as inseparable and are sampled together.

\section{Experimental Setup}

In this section, we detail the experimental setup for our audio-video and text-video experiments.
We describe the datasets, evaluations, and implementation details.
Section~\ref{sec:appendix-datasets} of the Appendix contains additional dataset details.

\subsection{Audio-Video Experiments}

\textbf{Training.} We train AVLnet on the instructional YouTube videos from the HowTo100M~\cite{miech2019howto100m} dataset. 
The HowTo100M dataset provides video clip segmentations according to time intervals of each video's ASR transcript and captions each clip with the text from its transcript.
However, to reduce the amount of supervision in our method, we train AVLnet on the video and audio from randomly segmented clips.

\textbf{Audio-Image Retrieval Evaluation.} 
Since instructional videos and spoken captions of images both contain descriptive speech of visual scenes, learning from instructional videos could provide a relevant initialization for learning from images and spoken captions.
Therefore, we train AVLnet on HowTo100M videos and fine-tune it on images and spoken captions in the Places Audio Caption Dataset.
The dataset contains 400k images from the Places205 dataset~\cite{zhou2014learning} paired with 1,000 hours of unscripted spoken captions.
We evaluate the performance on audio to image and image to audio retrieval tasks.
Following the prior work, results are reported on the validation set.
We use the standard recall metrics R@1, R@5, and R@10.

\textbf{Audio-Video Retrieval Evaluation.} We fine-tune and evaluate our model on two instructional video datasets: YouCook2~\cite{zhou2018towards} and CrossTask~\cite{zhukov2019cross}.
While YouCook2 contains cooking videos, CrossTask contains a wider range of instructional videos.
We also fine-tune and evaluate on MSR-VTT~\cite{xu2016msr-vtt} which contains general YouTube videos.
We use the human-annotated clips defined in each dataset: 9,586 train clips and 3,350 validation clips for YouCook2, 18,067 train clips and 2,852 validation clips for CrossTask, and 6,783 train clips and 968 test clips for MSR-VTT.
Please refer to Section~\ref{sec:appendix-datasets} of the Appendix for more dataset details.

We evaluate our model on video clip retrieval (audio to video) and language retrieval (video to audio) tasks, which measure how well the model can retrieve content in one modality based on a query in the other modality.
This follows prior work on audio to video retrieval on YouCook2~\cite{boggust2019grounding}.
This procedure tests our model's capability for video search directly using audio and spoken queries, without needing to transcribe speech in the query to text.
We report results in the zero-shot, fine-tuned, and no-pretraining settings.
We use the standard recall metrics R@1, R@5, and R@10.

\textbf{Implementation Details.} In the AVLnet audio branch, the audio input is represented as a log Mel filterbank spectrogram.
We use a 16 kHz sampling rate, 25 ms Hamming window, 10 ms window stride, and 40 Mel filter bands.
For the 2D and 3D visual feature extractors, we use the pretrained models from PyTorch~\cite{paszke2019pytorch} and feature extraction implementation provided by Miech~et al.~\cite{miech2019howto100m}. 
When training AVLnet, we do not update the weights of the 2D and 3D feature extractors due to GPU memory limitations.
We use a batch of $N=128$ videos, and sample $M=32$ clips per video, each $t=10$ seconds long. 
We minimize the MMS loss with Adam~\cite{kingma2015adam} using a learning rate of $1\mathrm{e}{-3}$ and fix the margin hyperparameter $\delta = 0.001$.
We train each model on 2 V100 GPUs for 30 epochs, which takes approximately 2 days.
For fine-tuning on the variable length video clips in the YouCook2, CrossTask, and MSR-VTT datasets, we crop or pad the audio up to 50s in YouCook2 and CrossTask, and 30s for audio in MSR-VTT.

\subsection{Text-Video Experiments}
\textbf{Training.} We train AVLnet-Text-Tri and AVLnet-Text-Fused on the instructional YouTube videos from the HowTo100M~\cite{miech2019howto100m} dataset.
For these experiments, we use the video clips defined by the time intervals of each video's ASR transcript, and we use the ASR text as the caption.

\textbf{Text-Video Retrieval Evaluation.} We evaluate and fine-tune our models on the YouCook2~\cite{zhou2018towards}, MSR-VTT~\cite{xu2016msr-vtt}, and LSMDC~\cite{rohrbach2017movie} datasets.
Each dataset provides human-annotated video clip boundaries and text summaries of the clips (full dataset details are Section~\ref{sec:appendix-datasets} of the Appendix).
We evaluate our models on the video clip and language retrieval tasks, in which a language query (text or text and audio) is used to retrieve video and vice versa.
The previous results~\cite{miech2019howto100m, miech2020end, amrani2020noise} on these datasets mainly focus on text to video retrieval (denoted by T\rarrow V).
Some models~\cite{yu2018joint, liu2019use} also incorporate audio into the retrieval task, where the audio is considered jointly with the video (denoted by T\rarrow A+V).
To compare with the prior work in this setting, we use the AVLnet-Text-Tri model.
Since the model is trained with a loss that encourages all three modalities to project into a shared embedding space, we use the sum of the text-video and text-audio similarities to retrieve the most similar videos to a given text caption.
We also consider the setting where audio is integrated with text and both are used to retrieve visual clips (denoted by T+A\rarrow V).
For this evaluation, we use the AVLnet-Text-Fused model. 
It is also possible to use the AVLnet-Text-Tri model for this evaluation, however, we found that that it typically performed worse than AVLnet-Text-Fused in this setting.
We use the standard recall metrics R@1, R@5, R@10, and the median rank (Md. R).

\textbf{Implementation Details.} For AVLnet-Text-Tri, the hyperparameters are the same as AVLnet, except we increased the batch size to 256, changed the learning rate to $2.5\mathrm{e}{-4}$, used a larger embedding size of $6144$, and used a clip length of $8$ seconds instead of $10$ seconds.
For AVLnet-Text-Fused, the hyperparameters are also the same as AVLnet, except we used a smaller batch size of 64 and smaller learning rate of $1\mathrm{e}{-4}$.
We trained both models for 15 epochs, using 4 V100 GPUs for AVLnet-Text-Tri and 2 V100 GPUs for AVLnet-Text-Fused.
\section{Audio-Video Experiments}

\begin{table*}[t]
    \begin{center}
    \caption{Video clip and language retrieval results on YouCook2, CrossTask, and MSR-VTT. Models trained on: (1) target dataset only (no pretraining); (2) HowTo100M only (zero-shot); (3) HowTo100M and target dataset (pretrain and fine-tune). All models use pretrained visual features.}
    \label{tab:retrieval}
    \begin{subtable}{0.9\linewidth}
        \vspace{-0.2cm}
        \caption{Video clip retrieval (A\rarrow V).}
        \label{tab:retrieval-video-clip}
        \vspace{-0.2cm}
        \resizebox{\linewidth}{!}{\begin{tabular}{l|rrr|rrr|rrr}
        \toprule
        \multicolumn{1}{l|}{\multirow{2}{*}{\textbf{Method (A\rarrow V)}}} & \multicolumn{3}{c|}{\textbf{YouCook2}}  & \multicolumn{3}{c|}{\textbf{CrossTask}}
        & \multicolumn{3}{c}{\textbf{MSR-VTT}} \\
        \multicolumn{1}{c|}{}   & \multicolumn{1}{c}{R@1}  & \multicolumn{1}{c}{R@5}  & \multicolumn{1}{c|}{R@10} & \multicolumn{1}{c}{R@1}  & \multicolumn{1}{c}{R@5}  & \multicolumn{1}{c|}{R@10} & \multicolumn{1}{c}{R@1}  & \multicolumn{1}{c}{R@5}  & \multicolumn{1}{c}{R@10} \\
        \midrule
        Random & 0.03 & 0.15 & 0.3 & 0.04 & 0.18 & 0.35 & 0.1 & 0.5 & 1.0 \\ 
        \midrule
        (1) Boggust et al.~\cite{boggust2019grounding} & 0.5 & 2.1 & 3.4 & 0.4 & 1.9 & 3.7 & 1.0 & 3.8 & 7.1 \\ 
        (1) Arandjelović et al.~\cite{arandjelovic2018objects} & 0.3 & 1.9 & 3.3 & 0.4 & \textbf{2.5} & 4.1 & \textbf{1.3} & 4.3 & 8.2 \\ 
        (1) \textbf{AVLnet} & \textbf{0.7} & \textbf{2.3} & \textbf{3.9} & \textbf{0.7} & 2.4 & \textbf{4.6} & 0.9 & \textbf{5.0} & \textbf{9.0} \\  
        \midrule
        (2) Boggust et al.~\cite{boggust2019grounding}  & 6.8 & 22.4 & 31.8 & 5.5 & 18.7 & 28.3 & 7.6 & 21.1 & 28.3 \\ 
        (2) Arandjelović et al.~\cite{arandjelovic2018objects} & 13.6 & 31.7 & 41.8 & 7.3 & 19.5 & 27.2 & 12.6 & 26.3 & 33.7 \\ 
        (2) \textbf{AVLnet} & \textbf{27.4} & \textbf{51.6} & \textbf{61.5} & \textbf{11.9} & \textbf{29.4} & \textbf{37.9} & \textbf{17.8} & \textbf{35.5} & \textbf{43.6} \\  
        \midrule
        (3) Boggust et al.~\cite{boggust2019grounding} & 8.5 & 26.9 & 38.5 & 6.6 & 20.8 & 31.2 & 10.3 & 27.6 & 35.9 \\ 
        (3) Arandjelović et al.~\cite{arandjelovic2018objects} & 17.4	& 39.7 & 51.5 & 9.5 & 25.8 & 36.6 & 16.2 & 32.2 & 42.9 \\
        (3) \textbf{AVLnet} & \textbf{30.7} & \textbf{57.7} & \textbf{67.4} & \textbf{13.8} & \textbf{34.5} & \textbf{44.8} & \textbf{20.1} & \textbf{40.0} & \textbf{49.6} \\
        \bottomrule
        \end{tabular}}
    \end{subtable}

    \begin{subtable}{0.9\linewidth}
        \vspace{0.2cm}
        \caption{Language retrieval (V\rarrow A).}
        \label{tab:retrieval-language}
        \vspace{-0.2cm}
        \resizebox{\linewidth}{!}{\begin{tabular}{l|rrr|rrr|rrr}
            \toprule
        \multicolumn{1}{l|}{\multirow{2}{*}{\textbf{Method (V\rarrow A)}}} & \multicolumn{3}{c|}{\textbf{YouCook2}}  & \multicolumn{3}{c|}{\textbf{CrossTask}}
        & \multicolumn{3}{c}{\textbf{MSR-VTT}} \\
        \multicolumn{1}{c|}{}   & \multicolumn{1}{c}{R@1}  & \multicolumn{1}{c}{R@5}  & \multicolumn{1}{c|}{R@10} & \multicolumn{1}{c}{R@1}  & \multicolumn{1}{c}{R@5}  & \multicolumn{1}{c|}{R@10} & \multicolumn{1}{c}{R@1}  & \multicolumn{1}{c}{R@5}  & \multicolumn{1}{c}{R@10} \\
        \midrule
        Random & 0.03 & 0.15 & 0.3 & 0.04 & 0.18 & 0.35 & 0.1 & 0.5 & 1.0 \\ 
        \midrule
        (1) Boggust et al.~\cite{boggust2019grounding} & 0.6 & 2.2 & 3.7 & 0.6 & 2.8 & 5.7 & 1.8 & 4.5 & 8.1 \\ 
        (1) Arandjelović et al.~\cite{arandjelovic2018objects} & 0.5 & 2.0 & 3.7 & \textbf{0.7} & 4.5 & 9.8 & 0.3 & 2.5 & 6.6 \\ 
        (1) \textbf{AVLnet} & \textbf{0.8} & \textbf{3.0} & \textbf{4.9} & 0.5 & \textbf{5.2} & \textbf{11.0} & \textbf{0.8} & \textbf{4.6} & \textbf{8.1}\\  
        \midrule
        (2) Boggust et al.~\cite{boggust2019grounding} & 7.9 & 23.8 & 32.3 & 5.2 & 18.2 & 27.6 & 9.3 & 20.7 & 28.8 \\ 
        (2) Arandjelović et al.~\cite{arandjelovic2018objects}  & 12.9 & 33.0 & 42.4 & 7.5 & 19.4 & 27.2 & 11.9 & 25.9 & 34.7 \\ 
        (2) \textbf{AVLnet} & \textbf{27.3} & \textbf{51.2} & \textbf{60.8} & \textbf{10.8} & \textbf{27.3} & \textbf{35.7} & \textbf{17.2} & \textbf{26.6} & \textbf{46.6} \\  
        \midrule
        (3) Boggust et al.~\cite{boggust2019grounding} & 9.9 & 30.0 & 41.1 & 6.0 & 21.5 & 31.4 & 11.8 & 29.0 & 38.6\\ 
        (3) Arandjelović et al.~\cite{arandjelovic2018objects} & 19.0 & 43.4 & 53.9 & 11.1 & 28.9 & 40.7 & 15.4 & 34.9 & 45.0 \\
        (3) \textbf{AVLnet} & \textbf{33.0} & \textbf{58.9} & \textbf{68.4} & \textbf{15.5} & \textbf{37.0} & \textbf{52.9} & \textbf{22.0} & \textbf{41.4} & \textbf{50.3} \\
        \bottomrule
        \end{tabular}}
    \end{subtable}
    \vspace{-0.5cm}
    \end{center}
\end{table*}

\begin{table}[t]
    \begin{center}
    \caption{Retrieval on Places using 400k training set. {\textdaggerdbl}Results found in~\cite{harwath2018jointly}. {\textdagger}Obtained using official code. *Concurrent work.}
    \label{tab:places}
	\resizebox{0.7\columnwidth}{!}{
    \begin{subtable}{\linewidth}
        \resizebox{\linewidth}{!}{\begin{tabular}{lrrrrrr}
        \toprule
        \multicolumn{1}{l}{\multirow{2}{*}{\textbf{Method}}} & \multicolumn{3}{c}{\textbf{Audio to Image}} & \multicolumn{3}{c}{\textbf{Image to Audio}}\\
        \multicolumn{1}{c}{}    & \multicolumn{1}{c}{R@1}  & \multicolumn{1}{c}{R@5}  & \multicolumn{1}{c}{R@10} & \multicolumn{1}{c}{R@1}  & \multicolumn{1}{c}{R@5}  & \multicolumn{1}{c}{R@10} \\
        \midrule
        Random & 0.1 & 0.5 & 1.0 & 0.1 & 0.5 & 1.0 \\ 
        Harwath et al.~\cite{harwath2016unsupervised}\textdaggerdbl & 14.8 & 40.3 & 54.8 & 12.1 & 33.5 & 46.3 \\
        Harwath et al.~\cite{harwath2017learning}\textdaggerdbl & 16.1 & 40.4 & 56.4 & 13.0 & 37.8 & 54.2 \\
        DAVEnet~\cite{harwath2018jointly} & 20.0 & 46.9 & 60.4 & 12.7 & 37.5 & 52.8 \\
        ResDAVEnet~\cite{harwath2020jointly} & 27.6 & 58.4 & 71.6 & 21.8 & 55.1 & 69.0 \\
        ResDAVEnet-VQ~\cite{harwath2020learning}{\textdagger} & 34.9 & 70.2 & 79.4 & 32.7 & 65.6 & 77.0 \\
        MILAN~\cite{sanabria2021talk}* & \textbf{53.4} & \textbf{79.1} & 86.3 & \textbf{53.0} & \textbf{78.2} & \textbf{85.6} \\
        \textbf{Ours, AVLnet} & 44.8 & 76.9 & \textbf{86.4} & 42.8 & 76.2 & 84.8 \\ 
        \bottomrule
        \end{tabular}}
    \end{subtable}}
    \end{center}
    \vspace{-0.5cm}
\end{table}

\subsection{Comparison to State-of-the-art}
\noindent \textbf{Audio-Image Retrieval.}
\label{sec:retrieval-results}
In this experiment, we train AVLnet on HowTo100M using the 2D CNN features so that it can be fine-tuned on the downstream images without any modifications.
During fine-tuning on Places, we update the weights of the visual encoder instead of keeping it frozen as in training on HowTo100M.
In Table~\ref{tab:places}, we compare prior models trained only on Places-400k~\cite{harwath2016unsupervised,harwath2017learning,harwath2018jointly,harwath2020jointly,harwath2020learning} to AVLnet trained on HowTo100M and fine-tuned on Places.
Our method achieves large gains over prior results, showing AVLnet learns a relevant initialization that transfers to the images and captions in Places.
We also show the results of concurrent work~\cite{sanabria2021talk} achieving similar results with different audio features and pretraining datasets.

\noindent \textbf{Audio-Video Retrieval.}
We compare AVLnet to prior audio-video models proposed for video clip retrieval in non-instructional contexts.
The model from Boggust~et~al.~\cite{boggust2019grounding} only uses the center image frame from each video clip during training and inference.
The model from Arandjelović et al.~\cite{arandjelovic2018objects} is trained with a binary cross-entropy loss.
Compared with AVLnet, it does not use non-linear gating and uses an embedding dimension of 128 instead of 4096.
For fair comparison, we train all models on HowTo100M, and, since the prior models each use different visual and audio pipelines, we change them to work with our 2D/3D visual features and deep audio network.

Table~\ref{tab:retrieval} shows the retrieval results on YouCook2, CrossTask, and MSR-VTT in the no-pretraining, zero-shot, and fine-tuned settings.
The performances on video clip retrieval (Table~\ref{tab:retrieval-video-clip}, A\rarrow V) and language retrieval (Table~\ref{tab:retrieval-language}, V\rarrow A) are similar for the same target dataset.
When trained only on the target dataset, the models all perform comparably.
Training on HowTo100M significantly improves the performance in the zero-shot and fine-tuned settings, suggesting that large-scale pretraining is essential.
This is true across all datasets, including on YouCook2 and CrossTask which contain instructional videos similar in content to HowTo100M videos, and on MSR-VTT which contains general videos.
AVLnet outperforms the baseline models, especially in the zero-shot and fine-tuned settings, and achieves significant performance on all datasets regardless of the domain.
Comparing the instructional datasets, the numbers are lower on CrossTask which suggest that it is a more challenging dataset for retrieval, possibly since it contains more general instructional videos.

\begin{table}[t]
\begin{center}
    \caption{AVLnet ablation study video clip retrieval (R@10). YC=YouCook2; CT=CrossTask; ZS=zero-shot; FT=fine-tune.}
    \label{tab:ablation-AVLnet}
	\resizebox{0.7\columnwidth}{!}{
	\begin{tabular}[t]{llrrrr}
	    \toprule
 		 \textbf{Study} & \textbf{Configuration}& \textbf{YC-ZS} & \textbf{YC-FT} & \textbf{CT-ZS} & \textbf{CT-FT} \\
		 \midrule
 		 \multirow{3}{*}{\shortstack[l]{Projection\\ Heads}} & Linear & 44.2 & 53.0 & 28.4 & 35.7 \\
		 & Non-Linear & 47.8 & 57.6 & 30.6 & 38.4 \\
 	     & Gating & \textbf{54.3} & \textbf{63.0} & \textbf{33.0} & \textbf{43.6} \\
 	     \midrule
		 \multirow{5}{*}{\shortstack[l]{Loss\\ Function}} & MIL-NCE & 24.8 & 29.6 & 15.2 & 22.1 \\
		 & Max-Margin & 27.4 & 39.1 & 18.7 & 30.1 \\
		 & Binary Cross Entropy & 46.2 & 54.6 & 28.4 & 41.3 \\
		 & InfoNCE & 51.6 & 60.5 & 31.9 & 41.9 \\
		 & MMS & \textbf{54.3} & \textbf{63.0} & \textbf{33.0} & \textbf{43.6} \\
		 \midrule
		 \multirow{3}{*}{\shortstack[l]{Clip Sampling / \\Visual Features}} & 2D features only & 51.6 & 57.9 & 32.6 & 37.9 \\
 		 & ASR clips & \textbf{57.6} & 62.8 & \textbf{34.6} & \textbf{44.5} \\
 	     & AVLnet & 54.3 & \textbf{63.0} & 33.0 & 43.6 \\
 	     \midrule
		 \multirow{4}{*}{Clip Duration} & 2.5s & 23.1 & 46.1 & 20.6 & 36.4 \\
 		 & 5s & 41.2 & 55.2 & 30.2 & 41.4 \\
 	     & 10s& \textbf{54.3} & \textbf{63.0} & \textbf{33.0} & \textbf{43.6} \\
 	     & 20s & 40.9 & 52.6 & 24.5 & 35.3 \\
		\bottomrule
	\end{tabular}}
\vspace{-0.5cm}
\end{center}
\end{table}
\begin{table}[t]
    \begin{center}
    \caption{Speech vs. non-speech retrieval results (R@10) on the Speech-241 and Sounds-241 evaluation sets derived from YouCook2.}
    \label{tab:speech-and-sounds} 
    \resizebox{0.5\linewidth}{!}{\begin{tabular}{l|rr|rr}
    \toprule
     \multicolumn{1}{l}{\multirow{2}{*}{\textbf{Method}}} & \multicolumn{2}{|c|}{\textbf{Speech-241}} & \multicolumn{2}{c}{\textbf{Sounds-241}} \\
     \multicolumn{1}{l}{} & \multicolumn{1}{|r}{\textbf{A\rarrow V}} & \multicolumn{1}{r}{\textbf{V\rarrow A}} & \multicolumn{1}{|r}{\textbf{A\rarrow V}} & \multicolumn{1}{r}{\textbf{V\rarrow A}}\\
    \midrule
    AVLnet zero-shot & 88.0 & 88.0  & 32.4 & 33.6\\ 
    AVLnet fine-tuned & \textbf{92.5} & \textbf{91.7} & 44.0 & 46.8\\ 
    \bottomrule
    \end{tabular}}
    \end{center}
\vspace{-0.5cm}
\end{table}
\subsection{Ablation Studies}
\label{sec:regression-analysis}
We evaluate our design choices via ablation studies comparing each model's video clip retrieval on YouCook2 and CrossTask (Table~\ref{tab:ablation-AVLnet}).
Given the computational requirements of HowTo100M, we train for 15 epochs with a batch size of 64.

\textbf{Projection Heads.}
First, we compare projections and find non-linear feature gating outperforms both linear and non-linear projection heads~\cite{chen2020simple}.

\textbf{Loss Functions.}
Next, we evaluate loss functions.
MMS~\cite{ilharco2019large} outperforms MIL-NCE~\cite{miech2020end}, Binary Cross Entropy~\cite{arandjelovic2017look}, Max-Margin Ranking~\cite{miech2019howto100m}, and InfoNCE~\cite{oord2018representation}. For MIL-NCE, we defined neighbors as the nearest non-overlapping 10s clips. For InfoNCE, we used negative samples from both within the same video and others.
MIL-NCE, initially proposed for text-video models, performs the worst, suggesting loss functions designed for text may not transfer well to audio.

\textbf{Visual Features and Clip Selection.}
We also find AVLnet performs better when trained on both 2D and 3D visual features.
AVLnet performs similarly when trained on random vs. ASR-defined clips, indicating our approach reduces supervision while maintaining performance.

\textbf{Clip Length.} 
Finally, we assess HowTo100M clip length and find it has a large effect on retrieval performance.
While we propose 10s, speech-image models~\cite{harwath2018jointly,harwath2020jointly} use spoken captions that are typically 20s, and text-video models~\cite{miech2020end} use ASR-defined clips that average 4s.
We find 10s outperforms 2.5, 5, and 20s, suggesting short clips may not contain speech relevant to the visuals, whereas long clips may contain too many audio-visual concepts.

\subsection{Retrieving Speech versus Non-Speech Sounds}
To identify the audio cues AVLnet uses for retrieval, we investigate performance in the absence and presence of speech.
We create two distinct evaluation sets: one containing videos without speech and one with speech.
To assign videos to each set, we identify the number of words in each YouCook2 validation video clip via ASR~\cite{WatsonSTT}.
We create a new evaluation set, Sounds-241, containing the 241 clips without a detected word.
We randomly sample 241 clips with at least one word detected to create another evaluation set: Speech-241.
AVLnet achieves higher retrieval performance on Speech-241 (Table~\ref{tab:speech-and-sounds}), suggesting our model is particularly effective when speech is present and supporting its application to speech to video search.
The performance on Sounds-241 is far above chance ($4.1\%$), demonstrating AVLnet also detects relevant cues in natural sounds.

\begin{figure*}[t]
    \centering 
        \includegraphics[width=0.99\linewidth]{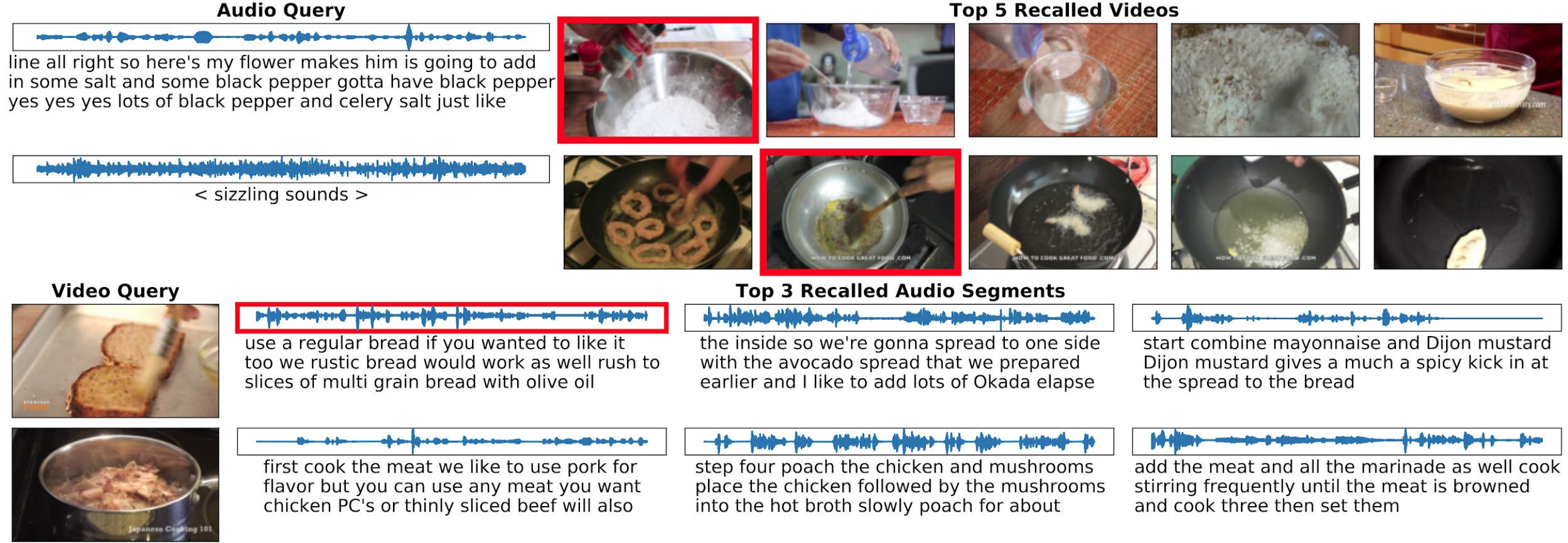}
    \caption{Video (top) and audio retrieval (bottom) results from AVLnet fine-tuned on YouCook2. Video clips are represented as their center frame, and audio clips are represented as their waveform and ASR transcript. The correct match is highlighted.}
    \label{fig:retrieval}
\end{figure*}
\subsection{Qualitative Retrieval Results}
\label{sec:avlnet-qualitative}
To better understand the performance gains AVLnet achieves over baseline methods, we analyze retrieval examples from our AVLnet model fine-tuned on YouCook2.
We show retrieval examples from the YouCook2 validation set in Figure~\ref{fig:retrieval}.
We find the retrieved results display high semantic similarity to salient content in the query.
For example, in the top row of Figure~\ref{fig:retrieval}, the query audio contains speech instructing viewers to mix together flour and other dry ingredients, and all the retrieved videos show bowls of flour mixtures.
The same is true for audio retrieval where, in the third row of Figure~\ref{fig:retrieval}, the query video clip shows oil spread on bread and the retrieved audio contains the words `bread' and `spread'.
This semantic relationship persists even when the correct clip is not the top result.
In the bottom row of Figure~\ref{fig:retrieval}, the correct clip is not recalled in the top five results, yet the video and retrieved audio are both related to cooking meat.
Further, we find AVLnet has learned to relate natural sounds to salient video clips.
The second row of Figure~\ref{fig:retrieval} shows an audio query containing only sizzling sounds.
Since there was no speech, the ASR system fails, but our model retrieves video clips of frying oil.
These results suggest our model has learned the semantic relationships between speech, natural sounds, and visual content, and support its application to video search directly using audio without transcribing speech.

\subsection{Audio-Visual Concept Discovery}
\label{sec:concept-discovery}

To understand the audio-visual concepts learned by our model, we apply unit visualization~\cite{zhou2015object} to AVLnet's multi-modal embedding space.
In this procedure, we rank dimensions of the latent space by the semantic similarity of their maximally activating audio and visual inputs.
By analyzing the consistency of the top dimensions, we can identify audio-visual concepts learned by our model. 

To compute the top dimensions, we begin by passing each YouCook2 validation clip through the AVLnet model fine-tuned on YouCook2.
Since the clips are up to a few minutes long, we remove the temporal pooling layer from the audio branch to get word-level activations.
Once we have a visual embedding and frame-level audio embedding for every clip, we identify the visual inputs and audio frames that maximally activate each dimension.
Each visual input is mapped to a set of visual food labels (provided by YouCook2~\cite{zhou2018weakly}), and each audio frame is mapped to a set of words from the ASR transcript during the 2 second window surrounding the frame.
Next, each dimension is given both a visual and audio label according to the most frequent [food label, word] in the dimension's top 50 most activating [visual, audio] inputs.
Using the visual and audio labels, we calculate each dimension's audio and visual purity as the fraction of the top 50 maximally activating visual or audio inputs that contain the correct label.
We sort the dimensions by the geometric mean of their purity scores and analyze the top dimensions.

The top 4 dimensions with unique labels are shown in Figure~\ref{fig:top_dimensions} (additional dimensions in Supplement).
Although the maximally activating visuals are chosen independently of the maximally activating audio, we find correspondences between the audio and visual content.
For example, dimension 201's audio and visual labels are `oil' and `pan' and its maximally activating clips show pans of oil.
This pattern continues in the other dimensions where we see strong correlations between the audio label, visual label, and the maximally activating clips, suggesting AVLnet has learned audio-visual concepts from raw instructional video.

\begin{figure}[t]
    \includegraphics[width=\linewidth]{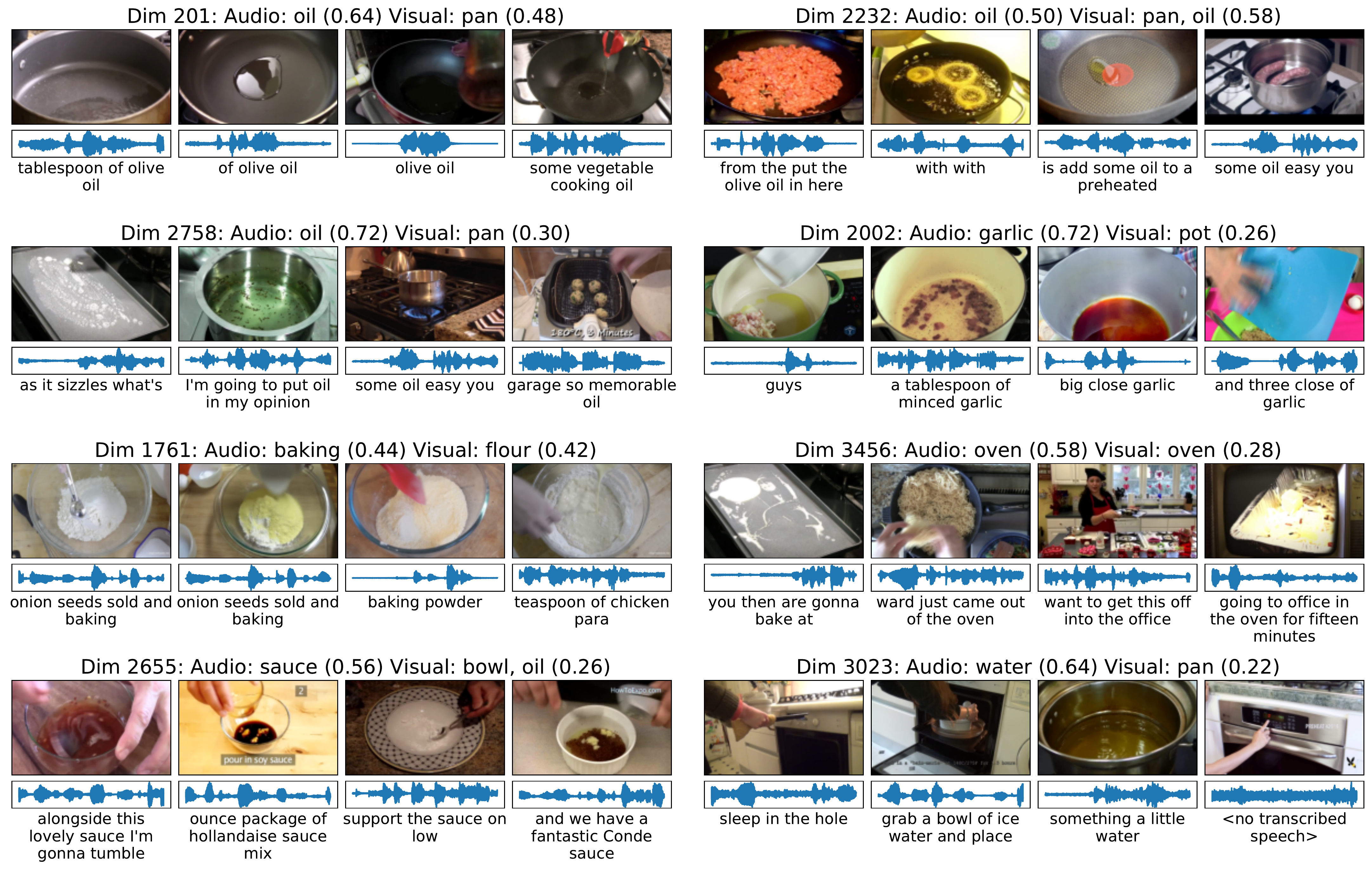}
    \caption{Top 8 dimensions sorted by geometric mean of their audio and visual purity displayed in row major order. Each dimension is represented as its top four visual features (shown as the clip's center frame) and top four frame-level audio features (shown as the frame's waveform and ASR transcript). The transcripts are shown for display purposes as AVLnet operates on video and raw audio.}
    \label{fig:top_dimensions}
\end{figure}
\section{Text-Video Experiments}
\subsection{Text-Video Retrieval Results}

\begin{table}[t]
    \begin{center}
    \caption{Text-Video retrieval results on YouCook2, MSR-VTT, and LSMDC. The best bi-modal and tri-modal results are bolded. Mod=Modalities.}
    \vspace{-0.20cm}
    \label{tbl:retrieval-avlnet-text}  
    \begin{subtable}{\linewidth}
        \caption{YouCook2}
        \vspace{-0.20cm}
        \resizebox{\linewidth}{!}{\begin{tabular}{llrrrrrlrrrr}
        \toprule
        \multicolumn{1}{l}{\multirow{2}{*}{\bf{Method}}} & \multicolumn{1}{l}{\multirow{2}{*}{\bf{Training Set}}} & \multicolumn{5}{c}{\bf{Video Clip Retrieval - YouCook2}}    & \multicolumn{5}{c}{\bf{Language Retrieval - YouCook2}}\\  
        \multicolumn{1}{c}{}    & \multicolumn{1}{c}{}     & \multicolumn{1}{c}{Mod.} & \multicolumn{1}{c}{R@1}  & \multicolumn{1}{c}{R@5}  & \multicolumn{1}{c}{R@10} & \multicolumn{1}{c}{Md. R} & \multicolumn{1}{c}{Mod.} & \multicolumn{1}{c}{R@1}  & \multicolumn{1}{c}{R@5}  & \multicolumn{1}{c}{R@10} & \multicolumn{1}{c}{Md. R} \\
        \midrule
        Random                & ---          &\rarrow V     & 0.03 & 0.15 & 0.3  & 1675 & V\rarrow       & 0.03 & 0.15 & 0.3  & 1675 \\ 
        Miech et al.~\cite{miech2019howto100m} & HT100M       & T\rarrow V   & 6.1  & 17.3 & 24.8 & 46   & V\rarrow T    & 5.3  & 16.5 & 25.2 & 42 \\ 
        Miech et al.~\cite{miech2020end} & HT100M       & T\rarrow V   & \bf{15.1} & \bf{38.0} & \bf{51.2} & \bf{10}   & ---            & ---  & ---  & ---  & --- \\
        Miech et al.~\cite{miech2019howto100m} & HT100M + YC2 & T\rarrow V   & 8.2  & 24.5 & 35.3 & 24   & V\rarrow T    & \bf{7.2}  & \bf{22.8} & \bf{34.3} & \bf{24}\\ \midrule \midrule
        
        AVLnet-Text-Tri   & HT100M       & T\rarrow A+V & 19.9 & 36.1 & 44.3 & 16.0     & V+A\rarrow T  & 28.5 & 53.7 & 65.3 & 6 \\ 
        AVLnet-Text-Tri   & HT100M + YC2 & T\rarrow A+V & \bf{30.2} & \bf{55.5} & \bf{66.5} & \bf{4}      & V+A\rarrow T & \bf{35.4} & \bf{63.3} & \bf{74.2} & \bf{4} \\
        
        \midrule \midrule
        AVLnet-Text-Fused   & HT100M       & T+A\rarrow V & 25.6 & 52.7 & 64.4 & 5    & V\rarrow T+A  & 29.3 & 55.3 & 65.5 & 4 \\ 
        AVLnet-Text-Fused   & HT100M + YC2 & T+A\rarrow V & \bf{33.2} & \bf{61.0} & \bf{71.5} & \bf{3}    & V\rarrow T+A  & \bf{34.0} & \bf{62.4} & \bf{72.5} & \bf{3} \\
        \bottomrule
        \end{tabular}}
    \label{tbl:youcook-retrieval-avlnet-text}
    \end{subtable}
    
    
    \begin{subtable}{\linewidth}
        \caption{MSR-VTT}
        \vspace{-0.20cm}
        \resizebox{\linewidth}{!}{\begin{tabular}{llrrrrrlrrrr}
        \toprule
        \multicolumn{1}{l}{\multirow{2}{*}{\bf{Method}}} & \multicolumn{1}{l}{\multirow{2}{*}{\bf{Training Set}}} & \multicolumn{5}{c}{\bf{Video Clip Retrieval - MSR-VTT}}    & \multicolumn{5}{c}{\bf{Language Retrieval - MSR-VTT}}\\  
        \multicolumn{1}{c}{}    & \multicolumn{1}{c}{}     & \multicolumn{1}{c}{Mod.} & \multicolumn{1}{c}{R@1}  & \multicolumn{1}{c}{R@5}  & \multicolumn{1}{c}{R@10} & \multicolumn{1}{c}{Md. R} & \multicolumn{1}{c}{Mod.} & \multicolumn{1}{c}{R@1}  & \multicolumn{1}{c}{R@5}  & \multicolumn{1}{c}{R@10} & \multicolumn{1}{c}{Md. R} \\
        \midrule
        Random                & ---          &\rarrow V     & 0.1 & 0.5 & 1.0  & 500 & V\rarrow       & 0.1 & 0.5 & 1.0  & 500 \\ 
        Miech et al.~\cite{miech2019howto100m} & HT100M       & T\rarrow V   & 7.5 & 21.2 & 29.6 & 38   & V\rarrow T    & 8.4 & 21.3 & 28.9 & 42 \\ 
        Amrani et al.~\cite{amrani2020noise}  & HT100M & T\rarrow V   & 8.0 & 21.3 & 29.3 & 33   & ---            & ---  & ---  & ---  & --- \\ 
        Miech et al.~\cite{miech2020end} & HT100M       & T\rarrow V   & 9.9 & 24.0 & 32.4 & 29.5   & ---            & ---  & ---  & ---  & --- \\ 
        Miech et al.~\cite{miech2019howto100m} & HT100M + MSR-VTT & T\rarrow V   & 14.9 & 40.2 & 52.8 & 9   & V\rarrow T    & \bf{16.8} & \bf{41.7} & \bf{55.1} & \bf{8} \\ 
        Amrani et al.~\cite{amrani2020noise} & HT100M + MSR-VTT & T\rarrow V   & \bf{17.4} & \bf{41.6} & \bf{53.6} & \bf{8}   & ---            & ---  & ---  & ---  & --- \\ \midrule \midrule
        
        JSFusion~\cite{yu2018joint}    & MSR-VTT          & T\rarrow A+V   & 10.2 & 31.2 & 43.2 & 13   & ---            & ---  & ---  & ---  & --- \\ 
        CE~\cite{liu2019use}  & MSR-VTT & T\rarrow A+V & 20.9 & 48.8 & 62.4 & 6 & V+A\rarrow T & 20.6 & 50.3 & \bf{64.0} & 5.3 \\ \midrule
        AVLnet-Text-Tri   & HT100M       & T\rarrow A+V & 8.3 & 19.2 & 27.4 & 47.5 & V+A\rarrow T & 8.7 & 19.6 & 25.1 & 45 \\ 
        AVLnet-Text-Tri   & HT100M + MSR-VTT & T\rarrow A+V & \bf{22.5} & \bf{50.5} & \bf{64.1} & \bf{5}  & V+A\rarrow T & \bf{22.5} & \bf{50.8} & 63.9 & \bf{5} \\ \midrule \midrule
        
        AVLnet-Text-Fused   & HT100M       & T+A\rarrow V & 19.6 & 40.8 & 50.7 & 9    & V\rarrow T+A  & 19.7 & 43.0 & 54.9 & 8 \\ 
        AVLnet-Text-Fused   & HT100M +  MSR-VTT & T+A\rarrow V & \bf{27.1} &  \bf{55.6} & \bf{66.6} & \bf{4}    & V\rarrow T+A  & \bf{28.5} &  \bf{54.6} & \bf{65.2} & \bf{4} \\
        \bottomrule
        \end{tabular}}
    \label{tbl:mrsvtt-retrieval-avlnet-text}
    \end{subtable}
    
    
    \begin{subtable}{\linewidth}
        \caption{LSMDC}
        \vspace{-0.20cm}
        \resizebox{\linewidth}{!}{\begin{tabular}{llrrrrrlrrrr}
        \toprule
        \multicolumn{1}{l}{\multirow{2}{*}{\bf{Method}}} & \multicolumn{1}{l}{\multirow{2}{*}{\bf{Training Set}}} & \multicolumn{5}{c}{\bf{Video Clip Retrieval - LSMDC}}    & \multicolumn{5}{c}{\bf{Language Retrieval - LSMDC}}\\  
        \multicolumn{1}{c}{}    & \multicolumn{1}{c}{}     & \multicolumn{1}{c}{Mod.} & \multicolumn{1}{c}{R@1}  & \multicolumn{1}{c}{R@5}  & \multicolumn{1}{c}{R@10} & \multicolumn{1}{c}{Md. R} & \multicolumn{1}{c}{Mod.} & \multicolumn{1}{c}{R@1}  & \multicolumn{1}{c}{R@5}  & \multicolumn{1}{c}{R@10} & \multicolumn{1}{c}{Md. R} \\
        \midrule
        Random                & ---          &\rarrow V     & 0.1 & 0.5 & 1.0  & 500 & V\rarrow       & 0.1 & 0.5 & 1.0  & 500 \\ 
        Miech et al.~\cite{miech2019howto100m} & HT100M       & T\rarrow V   & 4.0 & 9.8 & 14.0 & 137   & V\rarrow T    & 2.4 & 8.1 & 11.8 & 154 \\ 
        Amrani et al.~\cite{amrani2020noise} & HT100M & T\rarrow V   &  4.2 & 11.6 & 17.1 &  119 & ---            & ---  & ---  & ---  & --- \\
        Miech et al.~\cite{miech2019howto100m} & HT100M + LSMDC & T\rarrow V   & 7.1 & 19.6 & 27.9 & 40   & V\rarrow T    & \bf{6.6}   & \bf{17.8} & \bf{25.9} & \bf{50} \\ 
        Amrani et al.~\cite{amrani2020noise} & HT100M + LSMDC & T\rarrow V   & \bf{6.4} & \bf{19.8} & \bf{28.4} & \bf{39} & ---            & ---  & ---  & ---  & --- \\ \midrule \midrule
        
        JSFusion~\cite{yu2018joint}    & LSMDC & T\rarrow A+V   & 9.1  & 21.2 & 34.1 & 36   & ---  & ---  & ---  & ---  & --- \\ 
        CE~\cite{liu2019use}  & LSDMC & T\rarrow A+V & 11.2 & \bf{26.9} & \bf{34.8} & \bf{25.3} & --- & --- & --- & --- & --- \\ \midrule
        AVLnet-Text-Tri   & HT100M       & T\rarrow A+V & 1.4  & 5.9  & 9.4  & 273.5    & V+A\rarrow T  & 1.6 & 4.4 & 7.5 & 245.5 \\ 
        AVLnet-Text-Tri   & HT100M + LSDMC & T\rarrow A+V & \bf{11.4} & 26.0 & 34.6 & 30   & V+A\rarrow T  & \bf{12.1}  & \bf{25.5} & \bf{32.9} & \bf{34} \\ \midrule \midrule 
        
        AVLnet-Text-Fused   & HT100M       & T+A\rarrow V & 4.4 & 10.6 & 15.3 & 105.5    & V\rarrow T+A  & 3.8 & 11.3 & 15.9 & 109 \\ 
        AVLnet-Text-Fused   & HT100M +  LSMDC & T+A\rarrow V & \bf{17.0} &  \bf{38.0} & \bf{48.6} & \bf{11}    & V\rarrow T+A  & \bf{16.5}  & \bf{37.6} & \bf{47.6} & \bf{13} \\
        \bottomrule
        \end{tabular}}
    \label{tbl:lsmdc-retrieval-avlnet-text}
    \end{subtable}
\vspace{-0.5cm}
\end{center}
\end{table}

The retrieval results on YouCook2, MSR-VTT, and LSMDC are shown in Table~\ref{tbl:retrieval-avlnet-text}.
In general, the models that incorporate audio typically perform better than those that do not.
The improvement in performance when incorporating audio is more significant on YouCook2 and MSR-VTT than LSMDC, since the audio and visual channels in movies often have little salient alignment.
AVLnet-Text-Fused typically outperforms AVLnet-Text-Tri in terms of recall metrics on all datasets, but the retrieval setups differ (T+A\rarrow V versus T\rarrow A+V).
On YouCook2, both AVLnet-Text models outperform the previous state-of-the-art models; however, none of the previous models incorporated audio.
On MSR-VTT, AVLnet-Text-Tri outperforms the previous state-of-the-art that incorporated audio~\cite{liu2019use}.
On LSMDC, AVLnet-Text-Tri is on-par with the previous state-of-the-art model, achieving a higher R@1 result.

\subsection{Training with Text in a Low-Resource Scenario}
\label{sec:ablation-avlnet-text}
In this experiment, we explore a scenario where obtaining text annotations during training is expensive, but text exists or can be obtained for smaller evaluative datasets or real world applications.
We train the audio-video AVLnet model on HowTo100M without text, and fine-tune/evaluate it with the audio, video, and text from YouCook2, MSR-VTT, and LSMDC.
We integrate text into the model following the AVLnet-Text-Fused architecture design, and evaluate the model in the T+A\rarrow V setting. 
The results are shown in Table~\ref{tab:ablation-AVLnet-Text}, where we compare the zero-shot and fine-tuned results with AVLnet-Text-Fused.
Despite being trained on HowTo100M without any text and only fine-tuned with a small amount of text captions on the downstream datasets, the model can perform retrieval with text surprisingly well in both the zero-shot and fine-tuned conditions.
AVLnet-Text-Fused still achieves higher results, indicating that using ASR text captions during training on HowTo100M is beneficial.
Nonetheless, these results suggest that AVLnet learns language representations from speech, not just natural sounds or voice characteristics, and that audio representations can be adapted with text representations with only a small amount of text captions.

\begin{table}[t]
	\caption{Results on training with text in a low-resource scenario (R@10). Mod=Modalities, Eval=Evaluation, ZT=Zero-shot, FT=Fine-Tune.}
	    \centering
		\resizebox{0.7\columnwidth}{!}{\begin{tabular}[t]{ccrrcrrcrr}
		    \toprule
		    && \multicolumn{2}{c}{\textbf{YouCook2}} && \multicolumn{2}{c}{\textbf{MSR-VTT}} && \multicolumn{2}{c}{\textbf{LSMDC}}\\
			\textbf{HowTo100M Mod.} & \textbf{Eval. \& FT Mod.}& \textbf{ZT} & \textbf{FT} && \textbf{ZT} & \textbf{FT} && \textbf{ZT} & \textbf{FT} \\
			\midrule
		     A, V & T, A, V & 49.3 & 66.3 && 37.0 & 59.7 && 10.4 & 44.4\\
 			\midrule
			 T, A, V & T, A, V & \bf{64.4} & \bf{71.5} && \bf{50.7} & \bf{66.6} && \bf{15.3} & \bf{48.6}\\
		    \bottomrule
		\end{tabular}}
	\label{tab:ablation-AVLnet-Text}
\end{table}
\section{Conclusion}
\label{sec:conclusion}
In this paper, we present a self-supervised method for learning audio-video representations from instructional videos.
Whereas prior audio-video work mainly focuses on sound localization, our goal is to relate spoken words to visual entities.
We introduce the AVLnet model that learns directly from raw video, reducing the need for spoken or text annotations.
We establish baselines on video retrieval tasks on YouCook2, CrossTask, and MSR-VTT and achieve state-of-the-art performance on image retrieval tasks on the Places Spoken Caption dataset.
We show that AVLnet can learn audio-visual concepts by relating speech and sound to visual objects.
Further, we proposed a tri-modal model, AVLnet-Text, that additionally learns from the text narration which already exists in many instructional video datasets. 
The training method results in a multi-modal embedding space useful for text to video retrieval.
We plan to investigate the model's ability to learn representations in other languages as future work.

\section*{Broader Impact}
\label{sec:broader_impact}
In this paper, we introduced methods to learn correspondences between video and speech using video content naturally generated by humans instead of using manually annotated data.
This enables the possibility of learning correspondences in any language in the world with such video content. 
As less than 2\% of the world's languages have Automatic Speech Recognition (ASR) capability, this presents a significant opportunity.
Given the rapid adoption of video platforms by users globally, we expect that our methods could help scale the advancements in speech technologies developed for these languages.
This would enable a greater number of people to interact more effectively with computers.
\begin{ack} The authors are grateful for the support from the MIT-IBM Watson AI Lab. \end{ack}

\bibliographystyle{IEEEtranN}
{\small
\bibliography{ref}}

\begin{thebibliography}{75}
\providecommand{\natexlab}[1]{#1}
\providecommand{\url}[1]{#1}
\csname url@samestyle\endcsname
\providecommand{\newblock}{\relax}
\providecommand{\bibinfo}[2]{#2}
\providecommand{\BIBentrySTDinterwordspacing}{\spaceskip=0pt\relax}
\providecommand{\BIBentryALTinterwordstretchfactor}{4}
\providecommand{\BIBentryALTinterwordspacing}{\spaceskip=\fontdimen2\font plus
\BIBentryALTinterwordstretchfactor\fontdimen3\font minus
  \fontdimen4\font\relax}
\providecommand{\BIBforeignlanguage}[2]{{%
\expandafter\ifx\csname l@#1\endcsname\relax
\typeout{** WARNING: IEEEtranN.bst: No hyphenation pattern has been}%
\typeout{** loaded for the language `#1'. Using the pattern for}%
\typeout{** the default language instead.}%
\else
\language=\csname l@#1\endcsname
\fi
#2}}
\providecommand{\BIBdecl}{\relax}
\BIBdecl

\bibitem[Synnaeve et~al.(2014)Synnaeve, Versteegh, and
  Dupoux]{synnaeve2014learning}
G.~Synnaeve, M.~Versteegh, and E.~Dupoux, ``Learning words from images and
  speech,'' in \emph{NeurIPS Workshop on Learning Semantics}, 2014.

\bibitem[Harwath et~al.(2016)Harwath, Torralba, and
  Glass]{harwath2016unsupervised}
D.~Harwath, A.~Torralba, and J.~Glass, ``Unsupervised learning of spoken
  language with visual context,'' in \emph{NeurIPS}, 2016.

\bibitem[Harwath et~al.(2018)Harwath, Recasens, Sur{\'\i}s, Chuang, Torralba,
  and Glass]{harwath2018jointly}
D.~Harwath, A.~Recasens, D.~Sur{\'\i}s, G.~Chuang, A.~Torralba, and J.~Glass,
  ``Jointly discovering visual objects and spoken words from raw sensory
  input,'' in \emph{ECCV}, 2018.

\bibitem[Harwath et~al.(2020{\natexlab{a}})Harwath, Recasens, Sur{\'\i}s,
  Chuang, Torralba, and Glass]{harwath2020jointly}
------, ``Jointly discovering visual objects and spoken words from raw sensory
  input,'' in \emph{IJCV}, 2020.

\bibitem[Ilharco et~al.(2019)Ilharco, Zhang, and Baldridge]{ilharco2019large}
G.~Ilharco, Y.~Zhang, and J.~Baldridge, ``Large-scale representation learning
  from visually grounded untranscribed speech,'' in \emph{CoNLL}, 2019.

\bibitem[Chrupa{\l}a et~al.(2017)Chrupa{\l}a, Gelderloos, and
  Alishahi]{chrupala2017representations}
G.~Chrupa{\l}a, L.~Gelderloos, and A.~Alishahi, ``Representations of language
  in a model of visually grounded speech signal,'' in \emph{ACL}, 2017.

\bibitem[Merkx et~al.(2019)Merkx, Frank, and Ernestus]{merkx2019language}
D.~Merkx, S.~L. Frank, and M.~Ernestus, ``Language learning using speech to
  image retrieval,'' in \emph{INTERSPEECH}, 2019.

\bibitem[Sanabria et~al.(2021)Sanabria, Waters, and
  Baldridge]{sanabria2021talk}
R.~Sanabria, A.~Waters, and J.~Baldridge, ``Talk, don't write: A study of
  direct speech-based image retrieval,'' \emph{arXiv preprint
  arXiv:2104.01894}, 2021.

\bibitem[Aytar et~al.(2016)Aytar, Vondrick, and Torralba]{aytar2016soundnet}
Y.~Aytar, C.~Vondrick, and A.~Torralba, ``Soundnet: Learning sound
  representations from unlabeled video,'' in \emph{NeurIPS}, 2016.

\bibitem[Arandjelovic and Zisserman(2017)]{arandjelovic2017look}
R.~Arandjelovic and A.~Zisserman, ``Look, listen and learn,'' in \emph{ICCV},
  2017.

\bibitem[Arandjelovic and Zisserman(2018)]{arandjelovic2018objects}
------, ``Objects that sound,'' in \emph{ECCV}, 2018.

\bibitem[Owens and Efros(2018)]{owens2018audio}
A.~Owens and A.~A. Efros, ``Audio-visual scene analysis with self-supervised
  multisensory features,'' in \emph{ECCV}, 2018.

\bibitem[Korbar et~al.(2018)Korbar, Tran, and Torresani]{korbar2018cooperative}
B.~Korbar, D.~Tran, and L.~Torresani, ``Cooperative learning of audio and video
  models from self-supervised synchronization,'' in \emph{NeurIPS}, 2018.

\bibitem[Zhao et~al.(2018)Zhao, Gan, Rouditchenko, Vondrick, McDermott, and
  Torralba]{zhao2018sound}
H.~Zhao, C.~Gan, A.~Rouditchenko, C.~Vondrick, J.~McDermott, and A.~Torralba,
  ``The sound of pixels,'' in \emph{ECCV}, 2018.

\bibitem[Rouditchenko et~al.(2019)Rouditchenko, Zhao, Gan, McDermott, and
  Torralba]{rouditchenko2019self}
A.~Rouditchenko, H.~Zhao, C.~Gan, J.~McDermott, and A.~Torralba,
  ``Self-supervised audio-visual co-segmentation,'' in \emph{ICASSP}, 2019.

\bibitem[Miech et~al.(2018)Miech, Laptev, and Sivic]{miech2018learning}
A.~Miech, I.~Laptev, and J.~Sivic, ``Learning a text-video embedding from
  incomplete and heterogeneous data,'' \emph{arXiv preprint arXiv:1804.02516},
  2018.

\bibitem[Mithun et~al.(2018)Mithun, Li, Metze, and
  Roy-Chowdhury]{mithun2018learning}
N.~C. Mithun, J.~Li, F.~Metze, and A.~K. Roy-Chowdhury, ``Learning joint
  embedding with multimodal cues for cross-modal video-text retrieval,'' in
  \emph{ICMR}, 2018.

\bibitem[Wray et~al.(2019)Wray, Larlus, Csurka, and Damen]{wray2019fine}
M.~Wray, D.~Larlus, G.~Csurka, and D.~Damen, ``Fine-grained action retrieval
  through multiple parts-of-speech embeddings,'' in \emph{ICCV}, 2019.

\bibitem[Yu et~al.(2018)Yu, Kim, and Kim]{yu2018joint}
Y.~Yu, J.~Kim, and G.~Kim, ``A joint sequence fusion model for video question
  answering and retrieval,'' in \emph{ECCV}, 2018.

\bibitem[Liu et~al.(2019)Liu, Albanie, Nagrani, and Zisserman]{liu2019use}
Y.~Liu, S.~Albanie, A.~Nagrani, and A.~Zisserman, ``Use what you have: Video
  retrieval using representations from collaborative experts,'' \emph{arXiv
  preprint arXiv:1907.13487}, 2019.

\bibitem[Holzenberger et~al.(2019)Holzenberger, Palaskar, Madhyastha, Metze,
  and Arora]{holzenberger2019learning}
N.~Holzenberger, S.~Palaskar, P.~Madhyastha, F.~Metze, and R.~Arora, ``Learning
  from multiview correlations in open-domain videos,'' in \emph{ICASSP}, 2019.

\bibitem[Alayrac et~al.(2020)Alayrac, Recasens, Schneider, Arandjelovi{\'c},
  Ramapuram, De~Fauw, Smaira, Dieleman, and Zisserman]{alayrac2020self}
J.-B. Alayrac, A.~Recasens, R.~Schneider, R.~Arandjelovi{\'c}, J.~Ramapuram,
  J.~De~Fauw, L.~Smaira, S.~Dieleman, and A.~Zisserman, ``Self-supervised
  multimodal versatile networks,'' in \emph{NeurIPS}, 2020.

\bibitem[Zhou et~al.(2018{\natexlab{a}})Zhou, Xu, and Corso]{zhou2018towards}
L.~Zhou, C.~Xu, and J.~J. Corso, ``Towards automatic learning of procedures
  from web instructional videos,'' in \emph{AAAI}, 2018.

\bibitem[Miech et~al.(2020)Miech, Alayrac, Smaira, Laptev, Sivic, and
  Zisserman]{miech2020end}
A.~Miech, J.-B. Alayrac, L.~Smaira, I.~Laptev, J.~Sivic, and A.~Zisserman,
  ``End-to-end learning of visual representations from uncurated instructional
  videos,'' in \emph{CVPR}, 2020.

\bibitem[Miech et~al.(2019)Miech, Zhukov, Alayrac, Tapaswi, Laptev, and
  Sivic]{miech2019howto100m}
A.~Miech, D.~Zhukov, J.-B. Alayrac, M.~Tapaswi, I.~Laptev, and J.~Sivic,
  ``Howto100m: Learning a text-video embedding by watching hundred million
  narrated video clips,'' in \emph{ICCV}, 2019.

\bibitem[Sun et~al.(2019{\natexlab{a}})Sun, Baradel, Murphy, and
  Schmid]{sun2019learning}
C.~Sun, F.~Baradel, K.~Murphy, and C.~Schmid, ``Learning video representations
  using contrastive bidirectional transformer,'' \emph{arXiv preprint
  arXiv:1906.05743}, 2019.

\bibitem[Sun et~al.(2019{\natexlab{b}})Sun, Myers, Vondrick, Murphy, and
  Schmid]{sun2019videobert}
C.~Sun, A.~Myers, C.~Vondrick, K.~Murphy, and C.~Schmid, ``Videobert: A joint
  model for video and language representation learning,'' in \emph{ICCV}, 2019.

\bibitem[Sanabria et~al.(2018)Sanabria, Caglayan, Palaskar, Elliott, Barrault,
  Specia, and Metze]{sanabria18how2}
R.~Sanabria, O.~Caglayan, S.~Palaskar, D.~Elliott, L.~Barrault, L.~Specia, and
  F.~Metze, ``{How2:} a large-scale dataset for multimodal language
  understanding,'' in \emph{Workshop on Visually Grounded Interaction and
  Language (ViGIL)}.\hskip 1em plus 0.5em minus 0.4em\relax NeurIPS, 2018.

\bibitem[Prasad et~al.(2019)Prasad, van Esch, Ritchie, and
  Mortensen]{prasad2019building}
M.~Prasad, D.~van Esch, S.~Ritchie, and J.~F. Mortensen, ``Building
  large-vocabulary asr systems for languages without any audio training data.''
  in \emph{INTERSPEECH}, 2019.

\bibitem[Miech et~al.(2017)Miech, Laptev, and Sivic]{miech2017learnable}
A.~Miech, I.~Laptev, and J.~Sivic, ``Learnable pooling with context gating for
  video classification,'' in \emph{CVPR Workshop on YouTube-8M Large-Scale
  Video Understanding}, 2017.

\bibitem[Zhukov et~al.(2019)Zhukov, Alayrac, Cinbis, Fouhey, Laptev, and
  Sivic]{zhukov2019cross}
D.~Zhukov, J.-B. Alayrac, R.~G. Cinbis, D.~Fouhey, I.~Laptev, and J.~Sivic,
  ``Cross-task weakly supervised learning from instructional videos,'' in
  \emph{CVPR}, 2019.

\bibitem[Xu et~al.(2016)Xu, Mei, Yao, and Rui]{xu2016msr-vtt}
J.~Xu, T.~Mei, T.~Yao, and Y.~Rui, ``Msr-vtt: A large video description dataset
  for bridging video and language,'' in \emph{CVPR}, 2016.

\bibitem[Rohrbach et~al.(2017)Rohrbach, Torabi, Rohrbach, Tandon, Pal,
  Larochelle, Courville, and Schiele]{rohrbach2017movie}
A.~Rohrbach, A.~Torabi, M.~Rohrbach, N.~Tandon, C.~Pal, H.~Larochelle,
  A.~Courville, and B.~Schiele, ``Movie description,'' in \emph{IJCV}, 2017.

\bibitem[Harwath and Glass(2015)]{harwath2015deep}
D.~Harwath and J.~Glass, ``Deep multimodal semantic embeddings for speech and
  images,'' in \emph{ASRU}, 2015.

\bibitem[Harwath et~al.(2020{\natexlab{b}})Harwath, Hsu, and
  Glass]{harwath2020learning}
D.~Harwath, W.-N. Hsu, and J.~Glass, ``Learning hierarchical discrete
  linguistic units from visually-grounded speech,'' in \emph{ICLR}, 2020.

\bibitem[Suris et~al.(2019)Suris, Recasens, Bau, Harwath, Glass, and
  Torralba]{suris2019learning}
D.~Suris, A.~Recasens, D.~Bau, D.~Harwath, J.~Glass, and A.~Torralba,
  ``Learning words by drawing images,'' in \emph{CVPR}, 2019.

\bibitem[Mortazavi(2020)]{mortazavi2020speech}
M.~S. Mortazavi, ``Speech-image semantic alignment does not depend on any prior
  classification tasks,'' in \emph{INTERSPEECH}, 2020.

\bibitem[Wang et~al.(2021)Wang, Wang, Hasegawa-Johnson, Scharenborg, and
  Dehak]{wang2021align}
L.~Wang, X.~Wang, M.~Hasegawa-Johnson, O.~Scharenborg, and N.~Dehak, ``Align or
  attend? toward more efficient and accurate spoken word discovery using
  speech-to-image retrieval,'' in \emph{ICASSP}, 2021.

\bibitem[Zhou et~al.(2014)Zhou, Lapedriza, Xiao, Torralba, and
  Oliva]{zhou2014learning}
B.~Zhou, A.~Lapedriza, J.~Xiao, A.~Torralba, and A.~Oliva, ``Learning deep
  features for scene recognition using places database,'' in \emph{NeurIPS},
  2014.

\bibitem[Havard et~al.(2017)Havard, Besacier, and Rosec]{havard2017speech}
W.~Havard, L.~Besacier, and O.~Rosec, ``Speech-coco: 600k visually grounded
  spoken captions aligned to mscoco data set,'' \emph{arXiv preprint
  arXiv:1707.08435}, 2017.

\bibitem[Kamper and Roth(2018)]{kamper2018visually}
H.~Kamper and M.~Roth, ``Visually grounded cross-lingual keyword spotting in
  speech,'' \emph{arXiv preprint arXiv:1806.05030}, 2018.

\bibitem[Kamper et~al.(2019)Kamper, Anastassiou, and
  Livescu]{kamper2019semantic}
H.~Kamper, A.~Anastassiou, and K.~Livescu, ``Semantic query-by-example speech
  search using visual grounding,'' in \emph{ICASSP}, 2019.

\bibitem[Harwath and Glass(2017)]{harwath2017learning}
D.~Harwath and J.~Glass, ``Learning word-like units from joint audio-visual
  analysis,'' in \emph{ACL}, 2017.

\bibitem[Wang and Hasegawa-Johnson(2019)]{wang2019multimodal}
L.~Wang and M.~A. Hasegawa-Johnson, ``Multimodal word discovery and retrieval
  with phone sequence and image concepts.'' in \emph{INTERSPEECH}, 2019.

\bibitem[Wang and Hasegawa-Johnson(2020)]{wang2020dnn}
L.~Wang and M.~Hasegawa-Johnson, ``A dnn-hmm-dnn hybrid model for discovering
  word-like units from spoken captions and image regions,'' in
  \emph{INTERSPEECH}, 2020.

\bibitem[Leidal et~al.(2017)Leidal, Harwath, and Glass]{leidal2017learning}
K.~Leidal, D.~Harwath, and J.~Glass, ``Learning modality-invariant
  representations for speech and images,'' in \emph{ASRU}, 2017.

\bibitem[Eloff et~al.(2019)Eloff, Engelbrecht, and Kamper]{eloff2019multimodal}
R.~Eloff, H.~A. Engelbrecht, and H.~Kamper, ``Multimodal one-shot learning of
  speech and images,'' in \emph{ICASSP}, 2019.

\bibitem[Hsu and Glass(2018)]{hsu2018disentangling}
W.-N. Hsu and J.~Glass, ``Disentangling by partitioning: A representation
  learning framework for multimodal sensory data,'' \emph{arXiv preprint
  arXiv:1805.11264}, 2018.

\bibitem[Chrupa{\l}a(2021)]{chrupala2021visually}
G.~Chrupa{\l}a, ``Visually grounded models of spoken language: A survey of
  datasets, architectures and evaluation techniques,'' \emph{arXiv preprint
  arXiv:2104.13225}, 2021.

\bibitem[Monfort et~al.(2021)Monfort, Jin, Liu, Harwath, Feris, Glass, and
  Oliva]{monfort2021spoken}
M.~Monfort, S.~Jin, A.~Liu, D.~Harwath, R.~Feris, J.~Glass, and A.~Oliva,
  ``Spoken moments: Learning joint audio-visual representations from video
  descriptions,'' in \emph{CVPR}, 2021.

\bibitem[Oncescu et~al.(2020)Oncescu, Henriques, Liu, Zisserman, and
  Albanie]{oncescu2020queryd}
A.-M. Oncescu, J.~F. Henriques, Y.~Liu, A.~Zisserman, and S.~Albanie, ``Queryd:
  A video dataset with high-quality textual and audio narrations,'' \emph{arXiv
  preprint arXiv:2011.11071}, 2020.

\bibitem[Owens et~al.(2016)Owens, Wu, McDermott, Freeman, and
  Torralba]{owens2016ambient}
A.~Owens, J.~Wu, J.~H. McDermott, W.~T. Freeman, and A.~Torralba, ``Ambient
  sound provides supervision for visual learning,'' in \emph{ECCV}, 2016.

\bibitem[Hu et~al.(2019)Hu, Nie, and Li]{hu2019deep}
D.~Hu, F.~Nie, and X.~Li, ``Deep multimodal clustering for unsupervised
  audiovisual learning,'' in \emph{CVPR}, 2019.

\bibitem[Alwassel et~al.(2020)Alwassel, Mahajan, Korbar, Torresani, Ghanem, and
  Tran]{alwassel_2020_xdc}
H.~Alwassel, D.~Mahajan, B.~Korbar, L.~Torresani, B.~Ghanem, and D.~Tran,
  ``Self-supervised learning by cross-modal audio-video clustering,'' in
  \emph{NeurIPS}, 2020.

\bibitem[Gao et~al.(2018)Gao, Feris, and Grauman]{gao2018learning}
R.~Gao, R.~Feris, and K.~Grauman, ``Learning to separate object sounds by
  watching unlabeled video,'' in \emph{ECCV}, 2018.

\bibitem[Gao and Grauman(2019)]{gao20192}
R.~Gao and K.~Grauman, ``2.5d visual sound,'' in \emph{CVPR}, 2019.

\bibitem[Morgado et~al.(2018)Morgado, Nvasconcelos, Langlois, and
  Wang]{morgado2018self}
P.~Morgado, N.~Nvasconcelos, T.~Langlois, and O.~Wang, ``Self-supervised
  generation of spatial audio for 360 video,'' in \emph{NeurIPS}, 2018.

\bibitem[Yang et~al.(2020)Yang, Russell, and Salamon]{yang2020telling}
K.~Yang, B.~Russell, and J.~Salamon, ``Telling left from right: Learning
  spatial correspondence of sight and sound,'' in \emph{CVPR}, 2020.

\bibitem[Boggust et~al.(2019)Boggust, Audhkhasi, Joshi, Harwath, Thomas, Feris,
  Gutfreund, Zhang, Torralba, Picheny, and Glass]{boggust2019grounding}
A.~Boggust, K.~Audhkhasi, D.~Joshi, D.~Harwath, S.~Thomas, R.~Feris,
  D.~Gutfreund, Y.~Zhang, A.~Torralba, M.~Picheny, and J.~Glass, ``Grounding
  spoken words in unlabeled video,'' in \emph{CVPR Sight and Sound Workshop},
  2019.

\bibitem[Alayrac et~al.(2016)Alayrac, Bojanowski, Agrawal, Sivic, Laptev, and
  Lacoste-Julien]{alayrac2016unsupervised}
J.-B. Alayrac, P.~Bojanowski, N.~Agrawal, J.~Sivic, I.~Laptev, and
  S.~Lacoste-Julien, ``Unsupervised learning from narrated instruction
  videos,'' in \emph{CVPR}, 2016.

\bibitem[Tang et~al.(2019)Tang, Ding, Rao, Zheng, Zhang, Zhao, Lu, and
  Zhou]{Tang2019COIN}
Y.~Tang, D.~Ding, Y.~Rao, Y.~Zheng, D.~Zhang, L.~Zhao, J.~Lu, and J.~Zhou,
  ``Coin: A large-scale dataset for comprehensive instructional video
  analysis,'' in \emph{CVPR}, 2019.

\bibitem[Kuehne et~al.(2019)Kuehne, Iqbal, Richard, and Gall]{kuehne2019mining}
H.~Kuehne, A.~Iqbal, A.~Richard, and J.~Gall, ``Mining youtube-a dataset for
  learning fine-grained action concepts from webly supervised video data,''
  \emph{arXiv preprint arXiv:1906.01012}, 2019.

\bibitem[He et~al.(2016)He, Zhang, Ren, and Sun]{he2016deep}
K.~He, X.~Zhang, S.~Ren, and J.~Sun, ``Deep residual learning for image
  recognition,'' in \emph{CVPR}, 2016.

\bibitem[Deng et~al.(2009)Deng, Dong, Socher, Li, Li, and
  Fei-Fei]{deng2009imagenet}
J.~Deng, W.~Dong, R.~Socher, L.-J. Li, K.~Li, and L.~Fei-Fei, ``Imagenet: A
  large-scale hierarchical image database,'' in \emph{CVPR}, 2009.

\bibitem[Hara et~al.(2018)Hara, Kataoka, and Satoh]{hara2018can}
K.~Hara, H.~Kataoka, and Y.~Satoh, ``Can spatiotemporal 3d cnns retrace the
  history of 2d cnns and imagenet?'' in \emph{CVPR}, 2018.

\bibitem[Carreira and Zisserman(2017)]{carreira2017quo}
J.~Carreira and A.~Zisserman, ``Quo vadis, action recognition? a new model and
  the kinetics dataset,'' in \emph{CVPR}, 2017.

\bibitem[Oord et~al.(2018)Oord, Li, and Vinyals]{oord2018representation}
A.~v.~d. Oord, Y.~Li, and O.~Vinyals, ``Representation learning with
  contrastive predictive coding,'' \emph{arXiv preprint arXiv:1807.03748},
  2018.

\bibitem[Mikolov et~al.(2013)Mikolov, Chen, Corrado, and
  Dean]{mikolov2013efficient}
T.~Mikolov, K.~Chen, G.~Corrado, and J.~Dean, ``Efficient estimation of word
  representations in vector space,'' \emph{arXiv preprint arXiv:1301.3781},
  2013.

\bibitem[Paszke et~al.(2019)Paszke, Gross, Massa, Lerer, Bradbury, Chanan,
  Killeen, Lin, Gimelshein, Antiga, et~al.]{paszke2019pytorch}
A.~Paszke, S.~Gross, F.~Massa, A.~Lerer, J.~Bradbury, G.~Chanan, T.~Killeen,
  Z.~Lin, N.~Gimelshein, L.~Antiga \emph{et~al.}, ``Pytorch: An imperative
  style, high-performance deep learning library,'' in \emph{NeurIPS}, 2019.

\bibitem[Kingma and Ba(2015)]{kingma2015adam}
D.~P. Kingma and J.~Ba, ``Adam: A method for stochastic optimization,'' in
  \emph{ICLR}, 2015.

\bibitem[Amrani et~al.(2020)Amrani, Ben-Ari, Rotman, and
  Bronstein]{amrani2020noise}
E.~Amrani, R.~Ben-Ari, D.~Rotman, and A.~Bronstein, ``Noise estimation using
  density estimation for self-supervised multimodal learning,'' \emph{arXiv
  preprint arXiv:2003.03186}, 2020.

\bibitem[Chen et~al.(2020)Chen, Kornblith, Norouzi, and Hinton]{chen2020simple}
T.~Chen, S.~Kornblith, M.~Norouzi, and G.~Hinton, ``A simple framework for
  contrastive learning of visual representations,'' in \emph{ICML}, 2020.

\bibitem[Wat()]{WatsonSTT}
\url{https://www.ibm.com/watson/services/speech-to-text/}.

\bibitem[Zhou et~al.(2015)Zhou, Khosla, Lapedriza, Oliva, and
  Torralba]{zhou2015object}
B.~Zhou, A.~Khosla, A.~Lapedriza, A.~Oliva, and A.~Torralba, ``Object detectors
  emerge in deep scene cnns,'' in \emph{ICLR}, 2015.

\bibitem[Zhou et~al.(2018{\natexlab{b}})Zhou, Louis, and Corso]{zhou2018weakly}
L.~Zhou, N.~Louis, and J.~J. Corso, ``Weakly-supervised video object grounding
  from text by loss weighting and object interaction,'' in \emph{BMVC}, 2018.

\end{thebibliography}


\newpage
\appendix
\section*{Appendix}

\section{Illustration of the Masked Margin Softmax (MMS) Loss}
\label{sec:appendix-mms}

\begin{figure}[h]
    \centering 
        \includegraphics[width=0.7\linewidth]{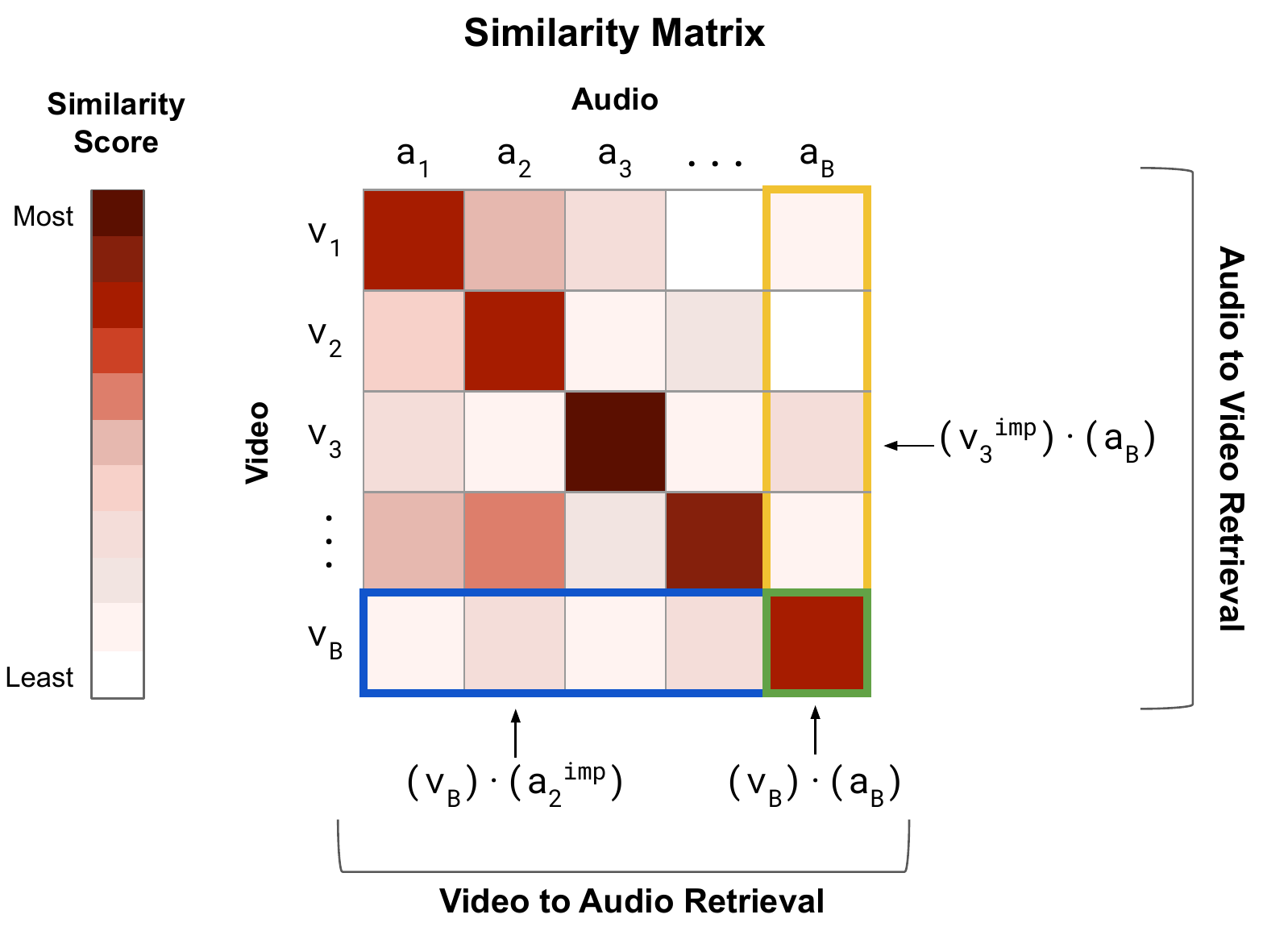}
    \caption{The MMS loss~\cite{ilharco2019large} maximizes the similarity of the true audio-visual pair $(\textbf{a}_i, \textbf{v}_i)$ shown in green. It also minimizes the similarity of $\textbf{a}_i$ paired with imposter videos $\textbf{v}_j^{\text{imp}}$ (in yellow) and $\textbf{v}_i$ paired with imposter audios $\textbf{a}_j^{\text{imp}}$ (in blue).}
    \label{fig:mms-loss}
\end{figure}

\section{Video Datasets}
\label{sec:appendix-datasets}
Many of our experiments are conducted on video datasets consisting of videos from YouTube.
These video datasets are typically distributed via lists of URLs to comply with YouTube's terms of service, and each research group must scrape the videos independently.
Over time, the number of videos available can shrink, making it challenging to reproduce and compare the results from different researchers.
Therefore, we provide details about the number of videos we were able to download, and which experimental splits we used.

\noindent \textbf{HowTo100M.}
The HowTo100M dataset~\cite{miech2019howto100m} contains instructional YouTube videos from domains such as \textit{home and garden}, \textit{computers and electronics} and \textit{food and entertaining}.
We downloaded the HowTo100M dataset from YouTube between Dec. 2019 - Mar. 2020.
The original dataset contains 1,238,792 videos.
At the time of download 1,166,089 videos were available on YouTube (72,703 less than the original dataset), which we used as our training set.

\noindent \textbf{YouCook2.}
The YouCook2 dataset~\cite{zhou2018towards} consists of 2,000 instructional cooking videos from YouTube.
The videos were separated into a 67-23-10 training-validation-testing split and categorized by humans into one of 89 recipe types (eg., \textit{spaghetti and meatballs}).
Videos were segmented by human annotators into clips representing recipe steps, and each clip was annotated with a text summary of the recipe step. 
Following Miech et al.~\cite{miech2019howto100m}, we use 9,586 training clips and 3,350 validation clips due to the unavailability of some videos on YouTube.

\noindent \textbf{CrossTask.}
The CrossTask dataset~\cite{zhukov2019cross} consists of 2,750 instructional videos from YouTube with 18 primary tasks and 65 related tasks. 
Each task is defined as list of steps, such as ``\textit{remove cap}'' and ``\textit{spread mixture}''.
Each video is associated with one task and contains a subset of steps from the task.
20 videos from each of the 18 primary tasks are designated as the validation set (360 videos total), and the remaining videos are designated as the training set.
The videos in the validation set were segmented into clips for each step by human annotators, while the videos in the training set were segmented into clips for each step automatically based on the ASR transcripts.
The training set contains 18,067 clips while the validation set contains 2,852 clips.

\noindent \textbf{MSR-VTT.}
The MSR-VTT~\cite{xu2016msr-vtt} dataset consists of YouTube videos from categories such as \textit{music} and \textit{sports} that are not necessarily instructional.
Videos were segmented into 10,000 video clips by human annotators and annotated with 20 natural language sentences each.
At the time of download, 5,722 videos with audio were available.
While several experimental splits have been proposed~\cite{liu2019use}, we use the \texttt{training-7k}~/~\texttt{test 1k-A} split, where the training set contains 7,000 video clips (proposed by~\cite{miech2019howto100m}) and the test set contains 1,000 video clips (proposed by~\cite{yu2018joint}).
Given that not all videos had audio, we train our model on 6,783 training clips and evaluate on 968 audio containing test clips.
For a fair comparison, we count the 32 missing test clips without audio as mistakes in our retrieval calculations.

\noindent \textbf{LSMDC.}
The LSMDC dataset~\cite{rohrbach2017movie} consists of movies with audio description (AD) --- audio descriptions of movie scenes for viewers with visual impairments.
The movies were split into video clips corresponding to scenes with AD narration, and each clip is annotated with the text transcript of the AD narration.
Following Miech et al.~\cite{miech2019howto100m}, we use 101,079 training clips and 1,000 testing clips.
We use the audio from the original movie clips; however, the audio is often silent because AD narration is inserted at breaks in dialogue.
The recorded AD narrations were not available. 

\section{Additional Concept Discovery Examples}
\label{sec:appendix-concept}
\begin{figure}[t]
    \centering
    \begin{subfigure}{0.7\linewidth}
        \centering
         \includegraphics[width=\linewidth]{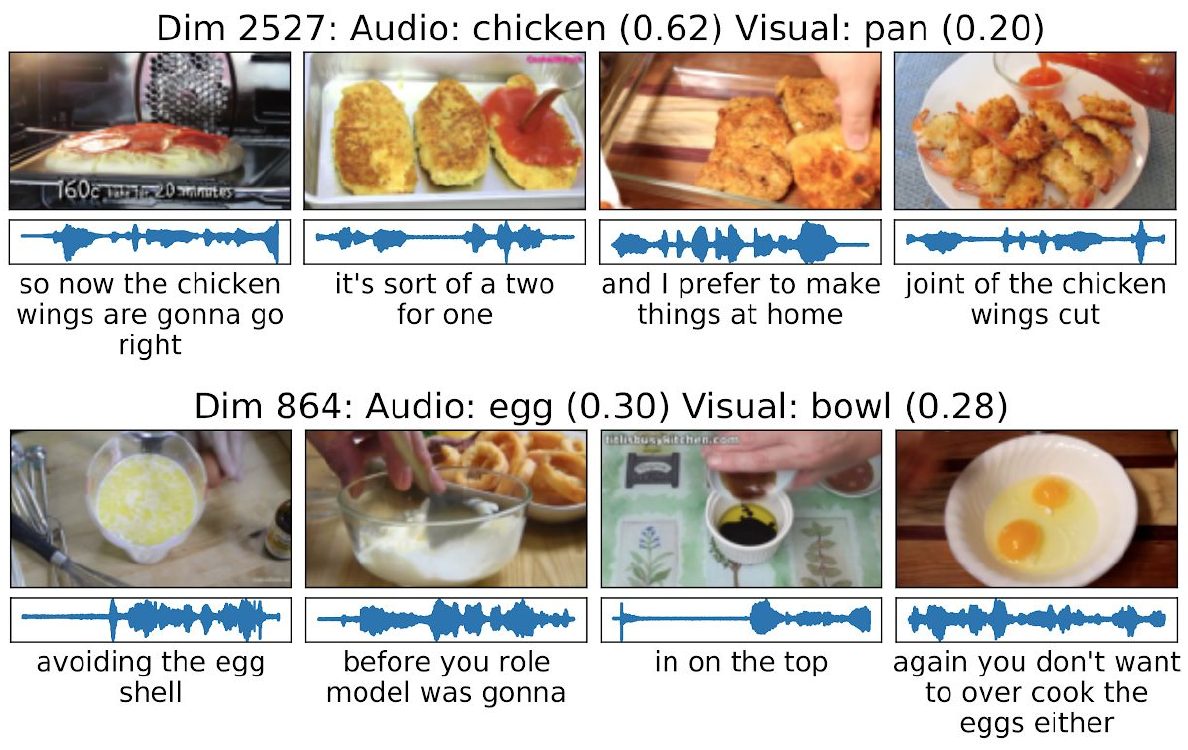}
         \vspace{-0.5cm} 
         \caption{Word activated dimensions.}  
         \vspace{0.3cm}
        \label{fig:supplement-word-dimensions}
    \end{subfigure}
    \vfill
    \begin{subfigure}{0.7\linewidth}
        \centering
         \includegraphics[width=\linewidth]{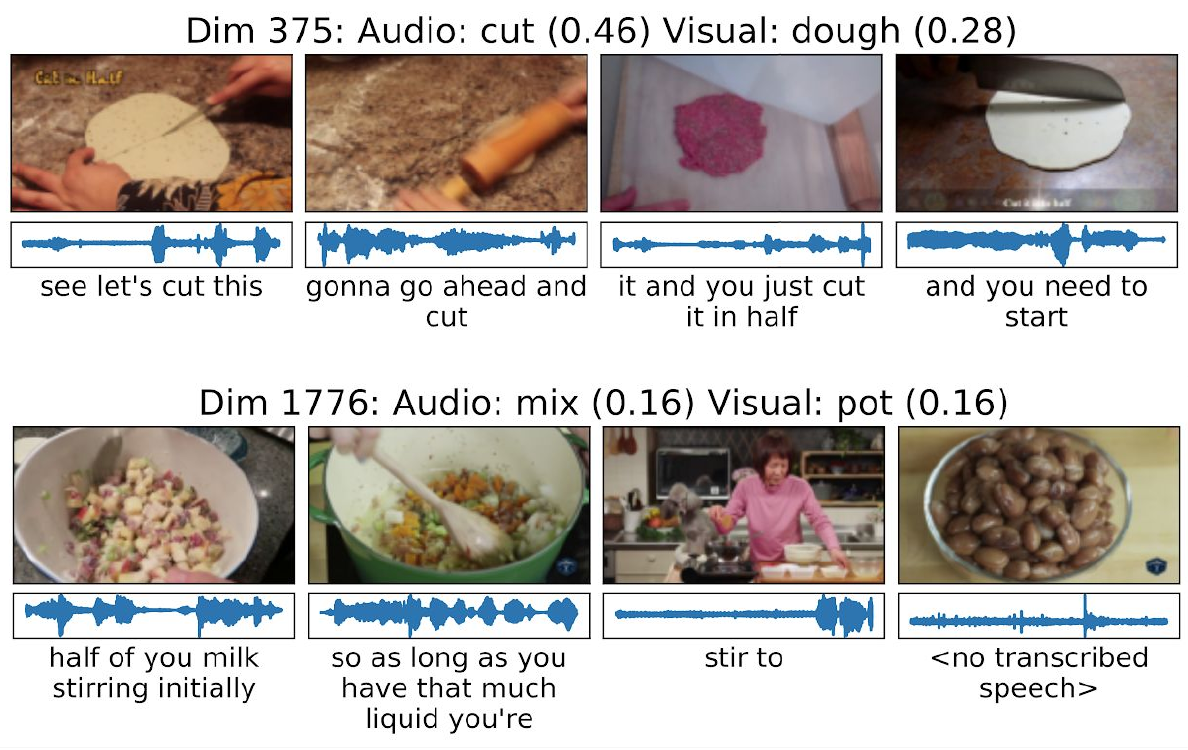}
         \vspace{-0.5cm}    
         \caption{Action activation dimensions.}
         \vspace{0.3cm}
        \label{fig:supplement-action-dimensions}
    \end{subfigure}
    \vfill
    \begin{subfigure}{0.7\linewidth}
        \centering
        \includegraphics[width=\linewidth]{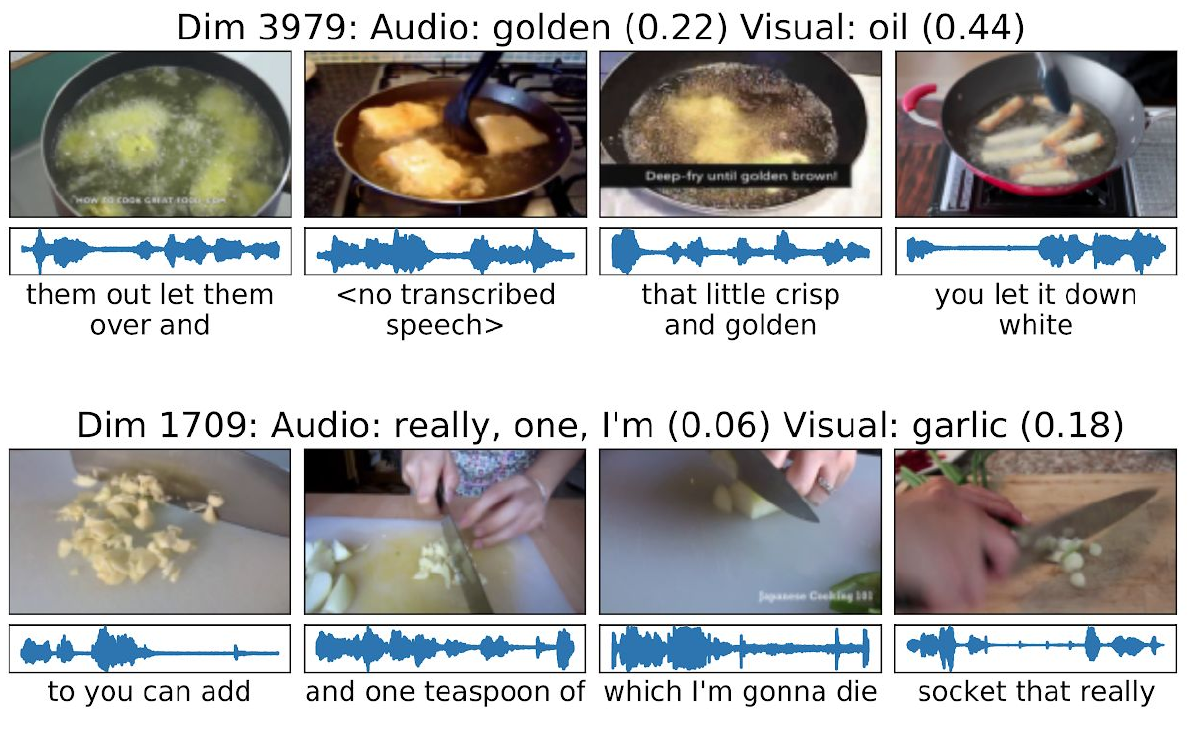}
        \vspace{-0.5cm} 
        \caption{Natural sound activated dimensions.}  
        \vspace{0.3cm}
        \label{fig:supplement-sound-dimensions}

    \end{subfigure}
\caption{Additional dimensions whose maximally activating audio and video features are semantically related. Each dimension is represented as its top four visual features (shown as the clip’s center frame) and top four frame-level audio features (shown as the frame’s waveform and ASR transcript). The transcripts are shown for display purposes as AVLnet operates on video and raw audio.}
\label{fig:supplement-dimensions}
\end{figure}

In Section~\ref{sec:concept-discovery}, we show AVLnet learns to relate semantically related audio and visual features to dimensions of the shared embedding space.
In Figure~\ref{fig:supplement-dimensions}, we show six additional dimensions that exhibit salient relationships between their maximally activating audio and visual segments.
In particular, Figure~\ref{fig:supplement-word-dimensions} shows dimensions that activate on words such as `chicken' and `egg' and Figure~\ref{fig:supplement-action-dimensions} shows dimensions that actions such as `cut' and `stir'.
In Figure~\ref{fig:supplement-sound-dimensions} we show dimensions that activate on natural sounds (ie. sizzling and chopping) as opposed to speech.

\section{Additional Video and Language Retrieval Examples}
\label{sec:appendix-retrieval}
\begin{figure*}[t]
    \begin{subfigure}{\linewidth}
         \includegraphics[width=\linewidth]{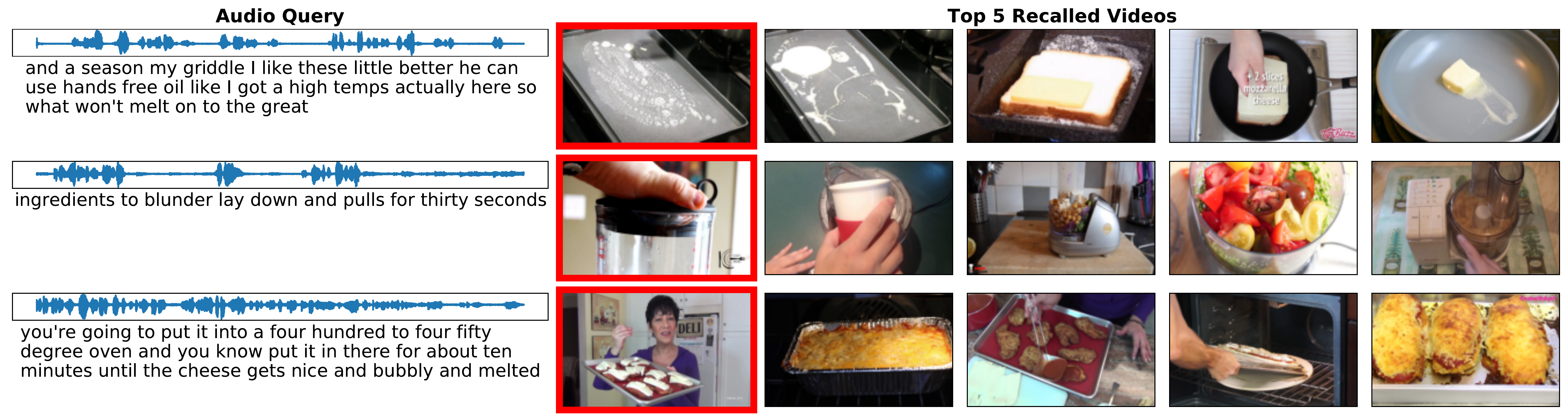}
          \caption{Video clip retrieval examples for clips retrieved correctly ($R@1$).}
        \vspace{0.3cm}
        \label{fig:supplement-youcook-r1-video-recall}
    \end{subfigure}
    \vfill
    \begin{subfigure}{\linewidth}
         \includegraphics[width=\linewidth]{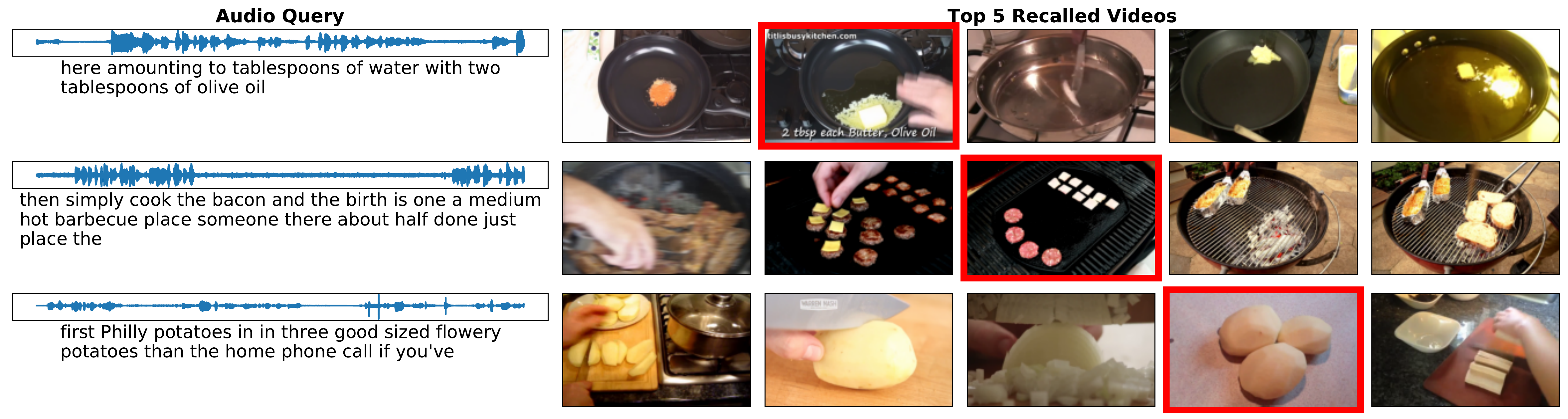}
          \caption{Video clip retrieval examples for clips retrieved in the top 5 results ($R@5$).}
         \vspace{0.3cm}
        \label{fig:supplement-youcook-r5-video-recall}
    \end{subfigure}
    \vfill
    \begin{subfigure}{\linewidth}
        \includegraphics[width=\linewidth]{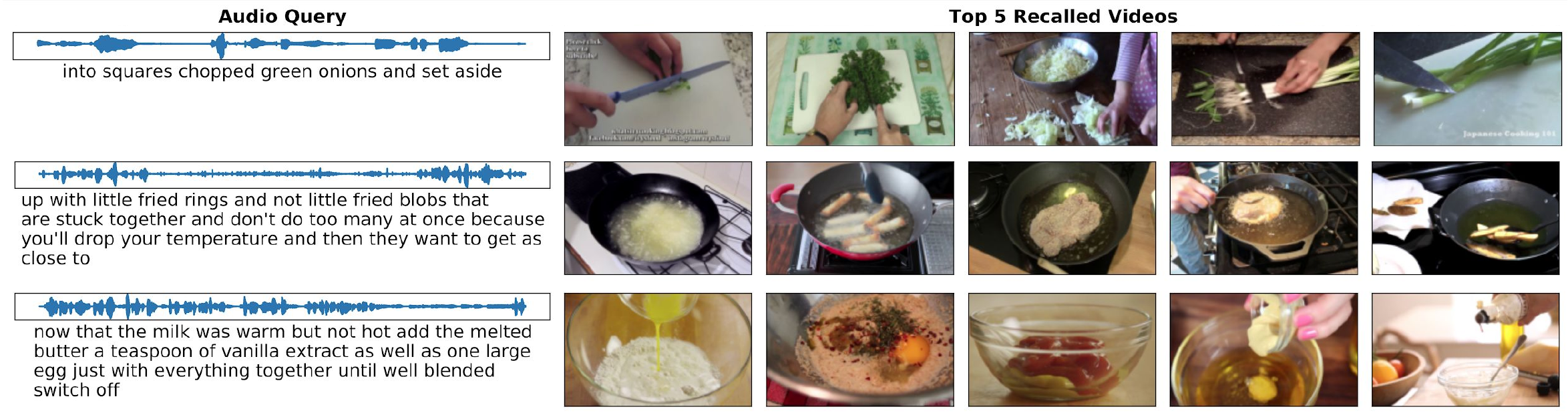}
        \caption{Video clip retrieval examples for clips not retrieved in the top 5 results ($R > 5$).}
         \vspace{0.3cm}
        \label{fig:supplement-youcook-r10-video-recall}
    \end{subfigure}
\caption{Additional video clip retrieval examples from the YouCook2 validation set. Each row displays the top recalled video clips (shown as each clip's center frame) to the given audio (shown as its waveform and ASR transcript). The ASR transcripts contain mistakes, but are only used for visualization given AVLnet operates on raw audio. The correct match is highlighted.}
\label{fig:supplement-youcook-video-recall}
\end{figure*}

\begin{figure*}[t]
    \begin{subfigure}{\linewidth}
         \includegraphics[width=\linewidth]{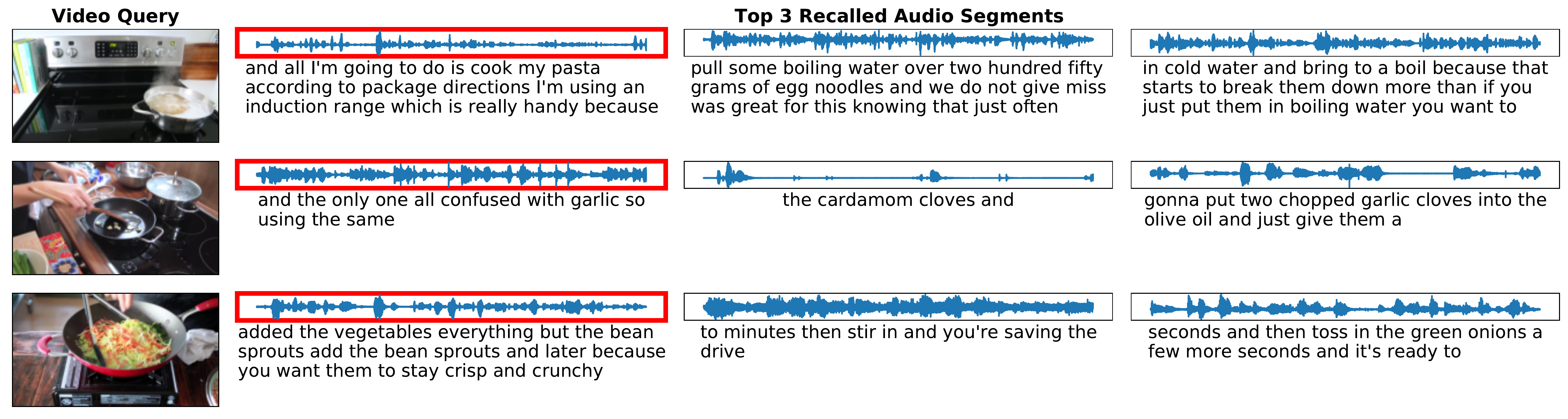}
        \caption{Language retrieval examples for clips retrieved correctly ($R@1$).}
         \vspace{0.3cm}
        \label{fig:supplement-youcook-r1-language-recall}
    \end{subfigure}
    \vfill
    \begin{subfigure}{\linewidth}
         \includegraphics[width=\linewidth]{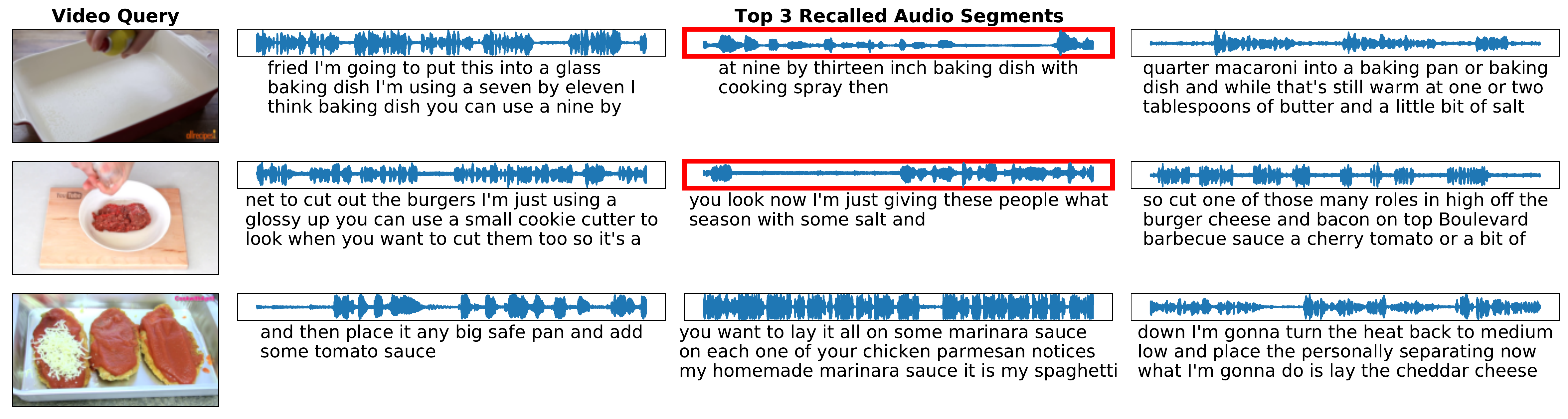}
        \caption{Language retrieval examples for clips retrieved in the top 5 results ($R@5$).}
         \vspace{0.3cm} 
        \label{fig:supplement-youcook-r5-language-recall}
    \end{subfigure}
    \vfill
    \begin{subfigure}{\linewidth}
        \includegraphics[width=\linewidth]{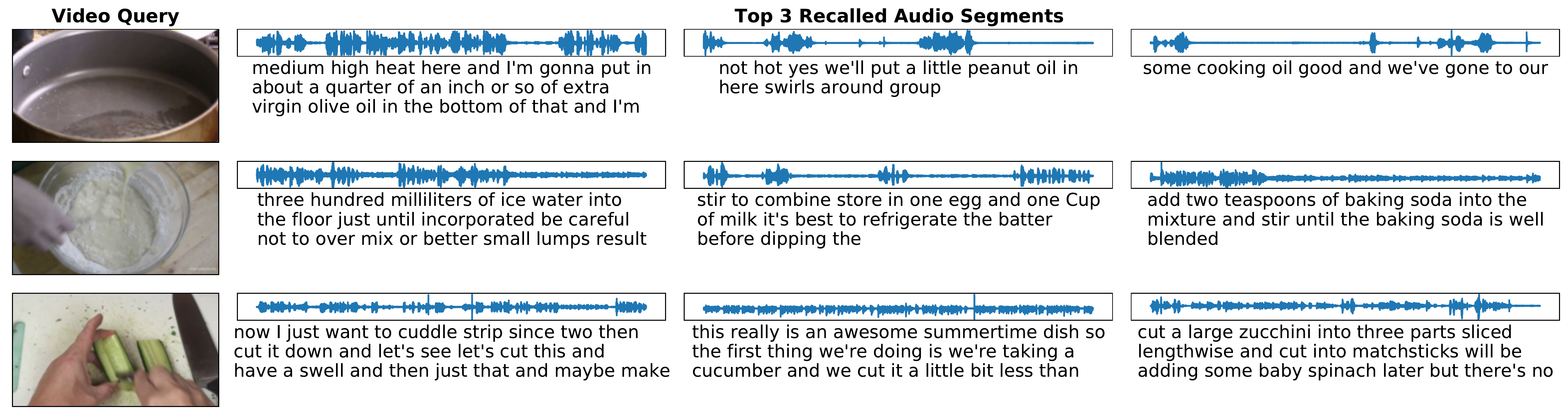}
        \caption{Language retrieval examples for clips not retrieved in the top 5 results ($R > 5$).}
         \vspace{0.3cm}
        \label{fig:supplement-youcook-r10-language-recall}
    \end{subfigure}
\caption{Additional audio retrieval examples from the YouCook2 validation set. Each row displays the top recalled audio segments (shown as each segment's waveform and ASR transcript) to the given video (shown as its center frame). The ASR transcripts contain mistakes, but are only used for visualization given AVLnet operates on raw audio. The correct match is highlighted.}
\label{fig:supplement-youcook-language-recall}
\end{figure*}

\begin{figure*}[h]
    \begin{subfigure}{\linewidth}
         \includegraphics[width=\linewidth]{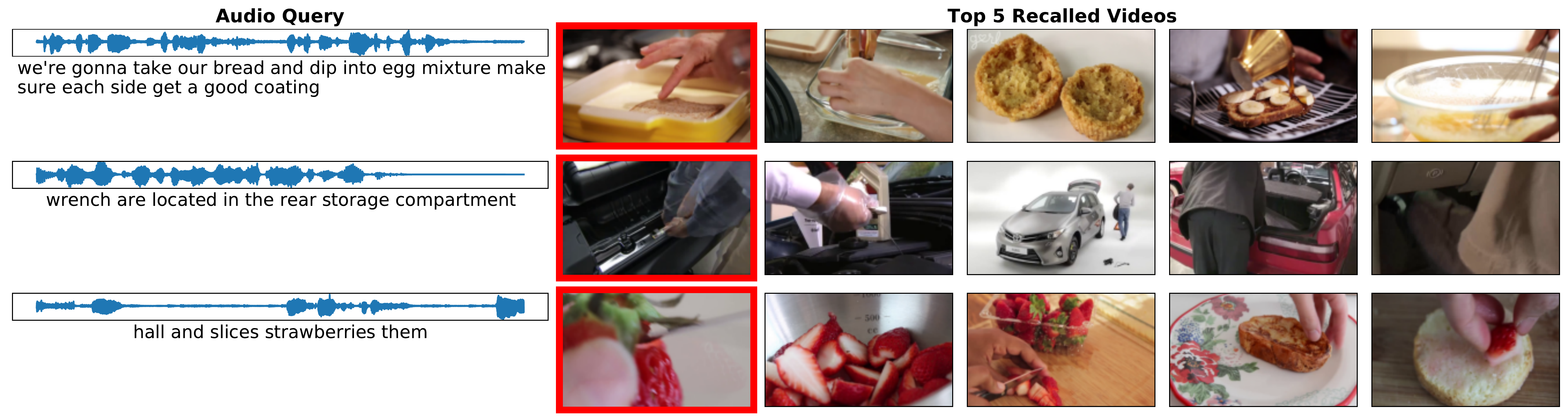}
        \caption{Video clip retrieval examples for clips retrieved correctly ($R@1$).}
         \vspace{0.3cm}
        \label{fig:supplement-crosstask-r1-video-recall}
    \end{subfigure}
    \vfill
    \begin{subfigure}{\linewidth}
         \includegraphics[width=\linewidth]{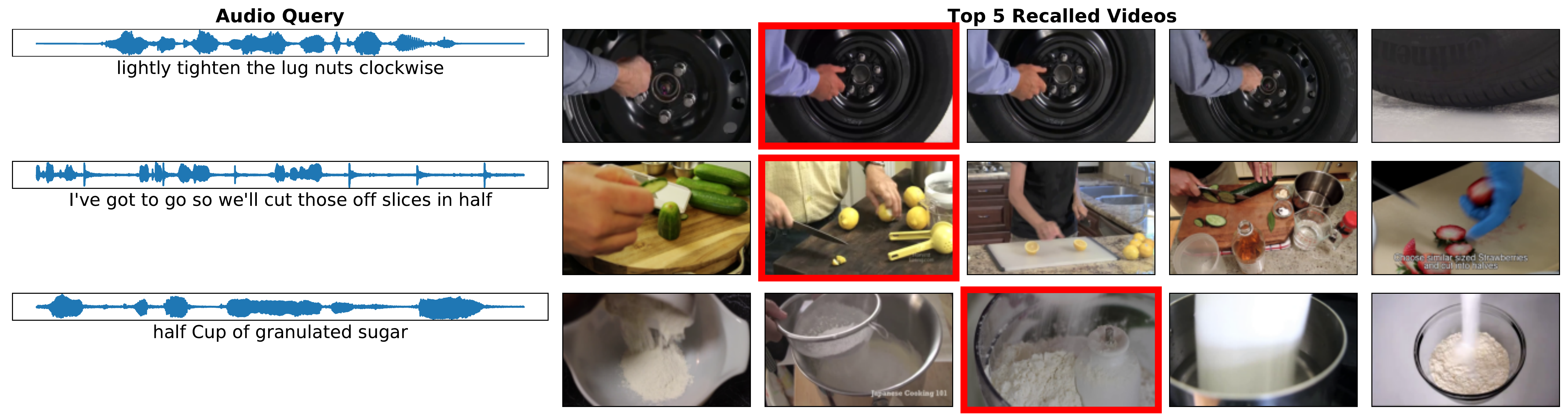}
        \caption{Video clip retrieval examples for clips retrieved in the top 5 results ($R@5$).}
         \vspace{0.3cm}
        \label{fig:supplement-crosstask-r5-video-recall}
    \end{subfigure}
    \vfill
    \begin{subfigure}{\linewidth}
        \includegraphics[width=\linewidth]{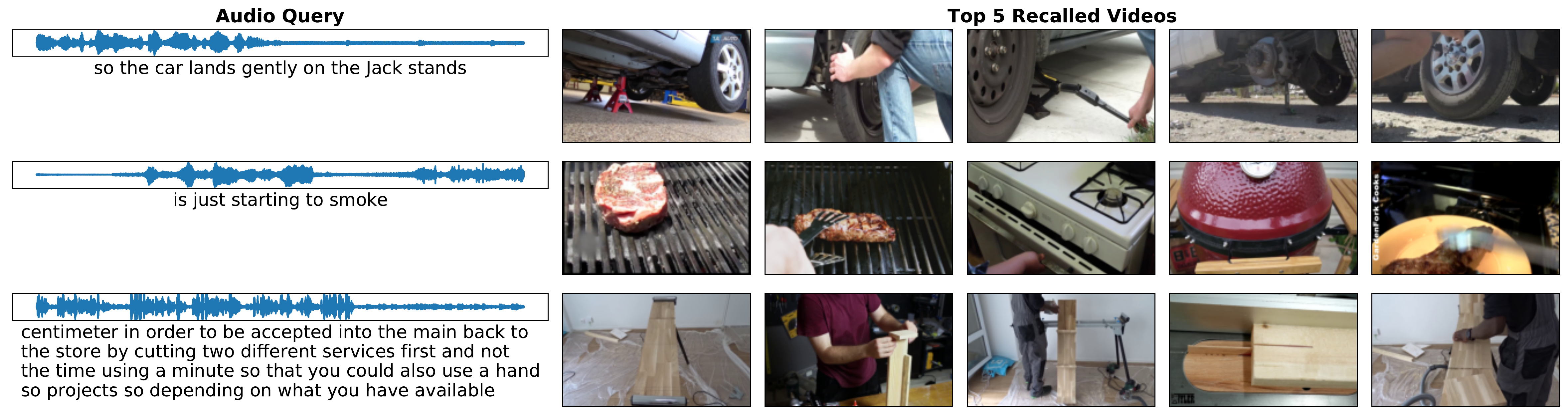}
        \caption{Video clip retrieval examples for clips not retrieved in the top 5 results ($R > 5$).}
         \vspace{0.3cm}
        \label{fig:supplement-crosstask-r10-video-recall}
    \end{subfigure}
\caption{Additional video clip retrieval examples from the CrossTask validation set. Each row displays the top recalled video clips (shown as each clip's center frame) to the given audio (shown as its waveform and ASR transcript). The ASR transcripts contain mistakes, but are only used for visualization given AVLnet operates on raw audio. The correct match is highlighted.}
\label{fig:supplement-crosstask-video-recall}
\end{figure*}

\begin{figure*}[h]
    \begin{subfigure}{\linewidth}
         \includegraphics[width=\linewidth]{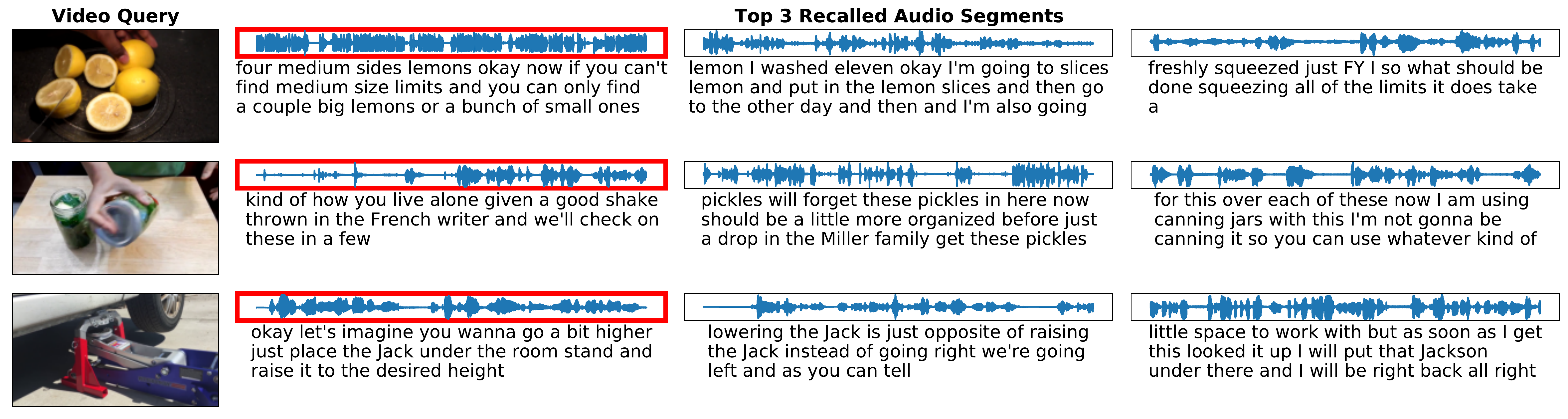}
        \caption{Audio retrieval examples for clips retrieved correctly ($R@1$).}
         \vspace{0.3cm} 
        \label{fig:supplement-crosstask-r1-language-recall}
    \end{subfigure}
    \vfill
    \begin{subfigure}{\linewidth}
         \includegraphics[width=\linewidth]{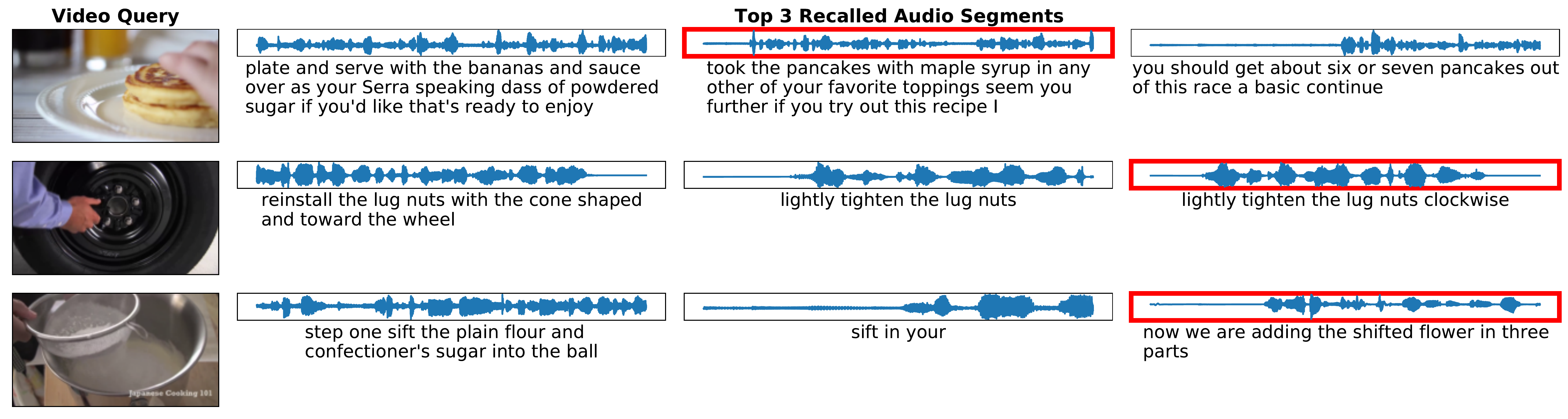}
        \caption{Audio retrieval examples for clips retrieved in the top 5 results ($R@5$).}
         \vspace{0.3cm}
        \label{fig:supplement-crosstask-r5-language-recall}
    \end{subfigure}
    \vfill
    \begin{subfigure}{\linewidth}
        \includegraphics[width=\linewidth]{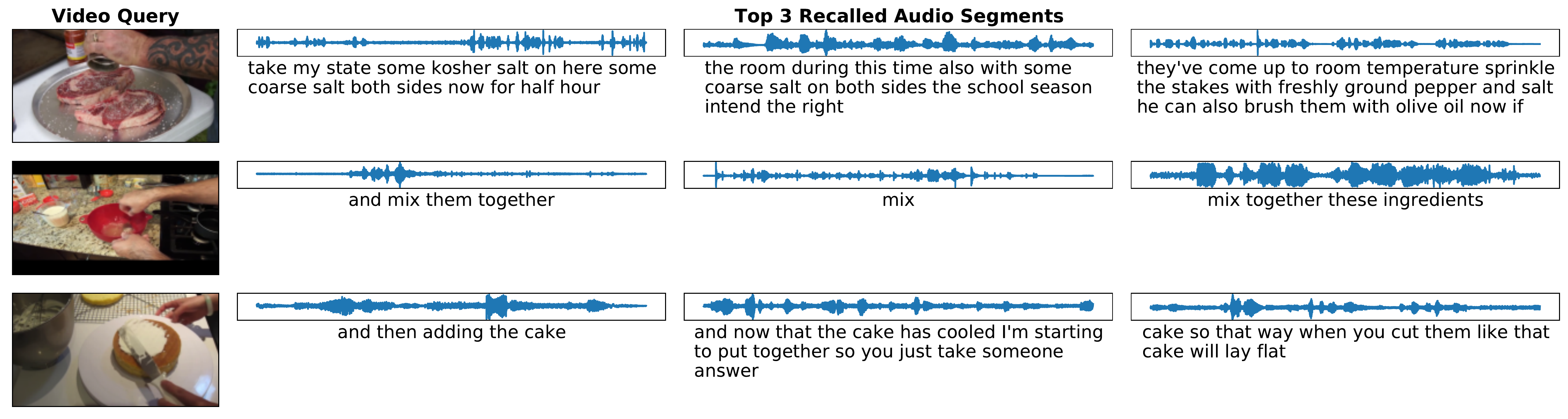}
        \caption{Audio retrieval examples for clips not retrieved in the top 5 results ($R > 5$).}
         \vspace{0.3cm}
        \label{fig:supplement-crosstask-r10-language-recall}
    \end{subfigure}
\caption{Additional audio retrieval examples from the CrossTask validation set. Each row displays the top recalled audio segments (shown as each segment's waveform and ASR transcript) to the given video (shown as its center frame). The ASR transcripts contain mistakes, but are only used for visualization given AVLnet operates on raw audio. The correct match is highlighted.}
\label{fig:supplement-crosstask-language-recall}
\end{figure*}

In Section~\ref{sec:retrieval-results}, we analyze the video and language retrieval results of our model and show qualitative retrieval examples in Figure~\ref{fig:retrieval}.
Here, we show additional video and audio retrieval examples from AVLnet fine-tuned on YouCook2 (Figures~\ref{fig:supplement-youcook-video-recall} and \ref{fig:supplement-youcook-language-recall}) and from AVLnet fine-tuned on CrossTask (Figures~\ref{fig:supplement-crosstask-video-recall} and \ref{fig:supplement-crosstask-language-recall}).
Consistent with our findings in Section~\ref{sec:avlnet-qualitative}, AVLnet retrieves clips that are semantically similar to the query clip, regardless of dataset.
In the YouCook2 examples, given an audio query instructing viewers to add ingredients to the blender (Figure~\ref{fig:supplement-youcook-r1-video-recall}) AVLnet recalls video clips of blenders, and given a video clip making hamburger patties (Figure~\ref{fig:supplement-youcook-r5-language-recall}) AVLnet recalls audio segments discussing burgers.
We find similar results on the CrossTask dataset where, given an audio query \textit{``lightly tighten the lug nuts clockwise''} (Figure~\ref{fig:supplement-crosstask-r5-video-recall}), AVLnet retrieves video clips tightening lug nuts on tires, and given a video query displaying cut lemons AVLnet retrieves audio segments about lemons (Figure~\ref{fig:supplement-crosstask-r1-language-recall}).
The similarity between queries and retrieved clips persists even when the correct result is not in AVLnet's top 5 results (Figures~\ref{fig:supplement-youcook-r10-video-recall}, \ref{fig:supplement-youcook-r10-language-recall}, \ref{fig:supplement-crosstask-r10-video-recall}, and \ref{fig:supplement-crosstask-r10-language-recall}).
For instance, in Figure~\ref{fig:supplement-youcook-r10-video-recall}, given an audio query about chopping green onions, AVLnet does not recall the correct clip in the top 5 results, but recalls other highly related clips of chopping green onions.
Overall, these results suggest AVLnet has learned to relate semantically similar audio and video channels of videos.


\end{document}